\documentclass[10pt,twocolumn,letterpaper]{article}

\usepackage[pagenumbers]{cvpr}

\definecolor{cvprblue}{rgb}{0.21,0.49,0.74}
\usepackage[pagebackref,breaklinks,colorlinks,allcolors=cvprblue]{hyperref}
\usepackage{bbm}
\usepackage{algorithm}
\usepackage{algorithmic}
\usepackage{booktabs}
\usepackage{multirow}
\usepackage{amsmath}
\usepackage{amssymb}
\usepackage{makecell}

\usepackage{graphicx}
\usepackage{subcaption}
\usepackage{pgfplots}
\usepackage{pgfplotstable}
\usepackage{tikz}

\newcolumntype{M}{>{\centering\arraybackslash}m{1.2cm}}

\title{BadPatch: Diffusion-Based Generation of Physical Adversarial Patches}

\author{
Zhixiang Wang\textsuperscript{1},    
Xingjun Ma\textsuperscript{1}\thanks{Corresponding author: {\tt\small xingjunma@fudan.edu.cn}},
Yu-Gang Jiang\textsuperscript{1} \\[0.5em]
\textsuperscript{1}Shanghai Key Lab of Intell. Info. Processing, School of CS, Fudan University 
}

\begin{document}
\maketitle
\begin{abstract}
Physical adversarial patches printed on clothing can enable individuals to evade person detectors, but most existing methods prioritize attack effectiveness over stealthiness, resulting in aesthetically unpleasing patches. While generative adversarial networks and diffusion models can produce more natural-looking patches, they often fail to balance stealthiness with attack effectiveness and lack flexibility for user customization.
To address these limitations, we propose \textbf{BadPatch}, a novel diffusion-based framework for generating customizable and naturalistic adversarial patches. Our approach allows users to start from a reference image (rather than random noise) and incorporates masks to create patches of various shapes, not limited to squares. To preserve the original semantics during the diffusion process, we employ Null-text inversion to map random noise samples to a single input image and generate patches through \textit{Incomplete Diffusion Optimization (IDO)}. Our method achieves attack performance comparable to state-of-the-art non-naturalistic patches while maintaining a natural appearance.
Using BadPatch, we construct \textbf{AdvT-shirt-1K}, the first physical adversarial T-shirt dataset comprising over a thousand images captured in diverse scenarios. AdvT-shirt-1K can serve as a useful dataset for training or testing future defense methods. The code and datasets are available at \url{https://github.com/Wwangb/BadPatch}.

\end{abstract}    
\section{Introduction}
\begin{figure}[!ht]
    \centering
    \includegraphics[width=0.85\columnwidth]{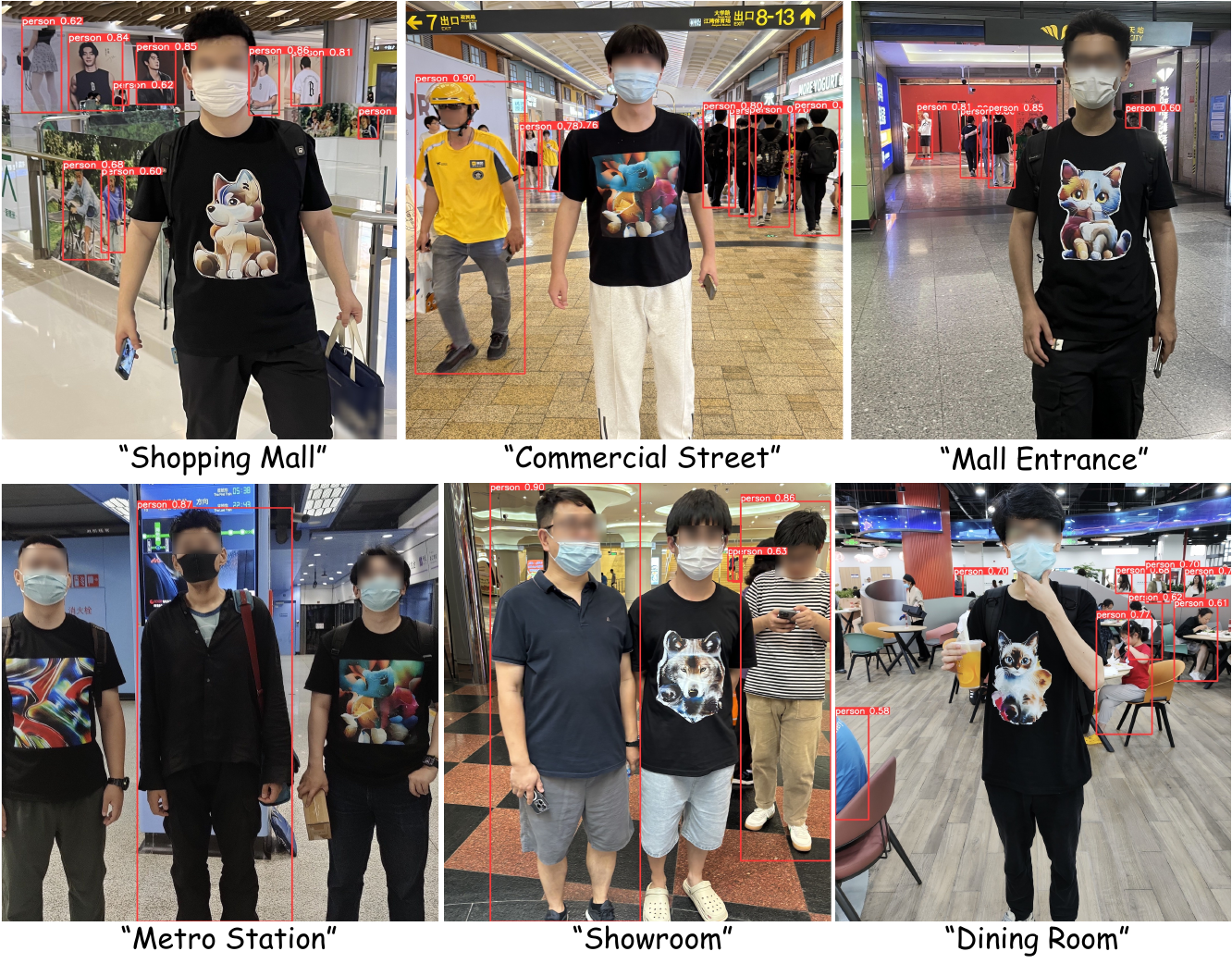}
    \caption{Example images from our \textbf{AdvT-shirt} dataset, showcasing the performance of BadPatch in various scenarios.}
    \label{fig:subset}
    \vspace{-0.5cm}
\end{figure}

Object detectors powered by deep neural networks (DNNs) are integral to modern computer vision systems, enabling critical applications such as autonomous driving \cite{chen2015deepdriving} and medical diagnostics \cite{miotto2018deep}. However, DNNs are highly susceptible to adversarial attacks \citep{carlini2017towards,xie2017adversarial,eykholt2018robust,wei2018transferable,ma2025safety,wang2025comprehensive}, which manipulate their outputs in ways that defy logic or compromise safety. These attacks can cause object detectors to misclassify or fail to detect critical objects. This vulnerability poses significant risks, particularly in safety-critical domains. 

Adversarial attacks on vision models can be broadly categorized into digital attacks \citep{goodfellow2014explaining,carlini2017towards} and physical attacks \citep{evtimov2017robust,kurakin2018adversarial,liu2022segment, li2024capgen, wei2025revisiting}. Unlike digital attacks, which introduce imperceptible perturbations in the digital domain, physical attacks manipulate real-world objects or environments to deceive vision models. Common physical attack strategies include applying adversarial patches \cite{thys2019fooling}, altering lighting conditions \cite{zhu2021fooling}, or introducing distracting elements \cite{liu2025beware}.

While adversarial patches have shown promise in physical attacks, most existing methods \citep{thys2019fooling,xu2020adversarial} prioritize attack effectiveness at the expense of stealthiness, resulting in patches that appear unnatural and are easily detectable. Although techniques like total variation loss \citep{wu2020making,huang2020universal} have been employed to smooth patches, they fail to fully address this issue. Recent advances in image generation \citep{goodfellow2014generative,ho2020denoising} have enabled the creation of more naturalistic patches \cite{hu2021naturalistic,lin2023diffusion}, but these methods often sacrifice attack effectiveness and lack customization capabilities. For instance, recent work \cite{chen2024content} leverages image-latent mapping for adversarial sample generation, but its large perturbations in physical settings lead to poor naturalness or effectiveness.

To address these limitations, we propose \textbf{BadPatch}, a novel adversarial patch generation framework based on Stable Diffusion \cite{rombach2022high}. BadPatch generates naturalistic and customizable adversarial patches from a reference image specified by the adversary. Our approach first maps the reference image into the latent space and optimizes noise latent vectors to create adversarial patches. We employ Null-text inversion \cite{mokady2023null} to ensure near-perfect reconstruction, preserving the original semantics. Additionally, we introduce an \emph{Incomplete Diffusion Optimization (IDO)} strategy to balance natural appearance and attack performance, along with an IoU-Detection loss to accelerate convergence. To generate irregular-shaped patches, we use masks to replace the background and suppress gradients in non-essential areas, reducing interference during optimization.

Using BadPatch, we construct \textbf{AdvT-shirt-1K}, a dataset of over a thousand images featuring customizable adversarial patches in diverse physical-world scenarios. This dataset demonstrates the practicality of our approach and provides a valuable resource for future research.

In summary, our key contributions are as follows:

\begin{itemize}
\item We propose a novel diffusion-based method called \textbf{BadPatch} for generating naturalistic adversarial patches based on a reference image. BadPatch integrates Null-text inversion, \emph{Incomplete Diffusion Optimization (IDO) }, masked customization, and an IoU-Detection loss to generate stylized and customizable adversarial patches.

\item Experimental results demonstrate that our BadPatch achieves the highest attack success rate (ASR), significantly reducing the mean average precision (mAP) of various object detectors. Its performance rivals or even surpasses state-of-the-art methods for generating unnatural adversarial patches.

\item We construct \textbf{AdvT-shirt-1K}, a dataset of physical-world adversarial patches printed on T-shirts. It includes 1,131 annotated images featuring 9 unique adversarial designs across diverse scenarios, including indoor and outdoor environments, as well as individual and group photos, providing a valuable resource for future defense research.
\end{itemize}
\section{Related Work}
\label{sec:related}


\subsection{Physical Adversarial Attacks}  
Early research on adversarial attacks primarily focused on digital domains, with methods like Basic Iterative Method (BIM) \cite{kurakin2018adversarial} and Projected Gradient Descent (PGD) \cite{madry2018towards} generating perturbations that remain effective even after physical transformations. To address real-world challenges, the Expectation Over Transformation (EOT) approach \cite{athalye2018synthesizing} was introduced, simulating environmental variations to enhance robustness. The Digital-to-Physical (D2P) method \cite{jan2019connecting} further improved attack effectiveness by modeling the transition between digital and physical domains. These advancements laid the groundwork for physical adversarial attacks, such as AdvPatch \cite{brown2017adversarial}, which demonstrated robust adversarial patches effective across diverse conditions. Subsequent work, including AdvYOLO \cite{thys2019fooling}, extended these ideas to evade object detectors using clothing-based patches. Other physical attack strategies have leveraged accessories like glasses \cite{sharif2016accessorize}, manipulated lighting \cite{zhu2021fooling}, and exploited shadows \cite{zhong2022shadows}, highlighting the expanding scope of real-world adversarial threats.

\subsection{Naturalistic Adversarial Patch}  
Drawing inspiration from the imperceptibility constraints in digital attacks, Universal Physical Camouflage Attacks (UPC) \cite{huang2020universal} use the $L_\infty$ norm to bound perturbations, ensuring patches retain a natural appearance. Unlike traditional iterative methods that optimize patches directly on images, the Naturalistic Adversarial Patch (NAP) \cite{hu2021naturalistic} adopts a generative approach. NAP employs GANs to optimize patches indirectly by modifying initial latent vectors, producing patches that are both natural-looking and effective. This approach has spurred the development of several GAN-based methods \citep{doan2022tnt,lapid2023patch}. More recently, diffusion models have emerged as a powerful tool in the vision domain \citep{chen2023natural,lin2023diffusion,xue2024diffusion}, leveraging their generative capabilities to create even more naturalistic and effective physical adversarial patches. These advancements highlight the growing potential of generative models in crafting stealthy and robust adversarial attacks.

\begin{figure*}[!ht]
    \centering
    \includegraphics[width=0.8\textwidth]{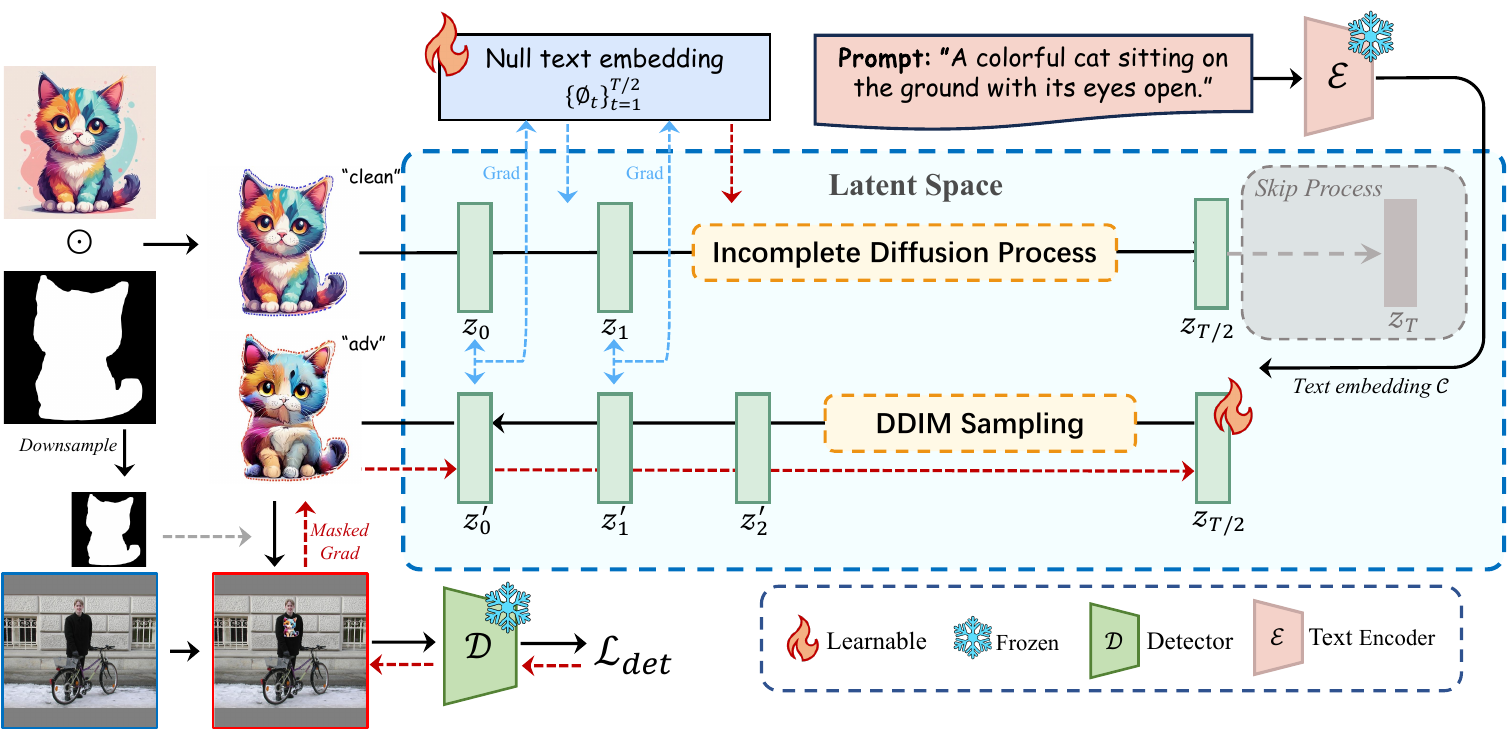}
    \caption{An overview of the proposed \emph{BadPatch} generation framework.}
    \label{fig:overview}
    \vspace{-0.4cm}
\end{figure*}

\subsection{Diffusion-based Image Editing}
Diffusion models have recently achieved remarkable success in image generation, with models like DALL-E2 \cite{ramesh2022hierarchical}, Imagen \cite{saharia2022photorealistic}, and Stable Diffusion \cite{rombach2022high} outperforming GANs in various benchmarks \cite{dhariwal2021diffusion}. Beyond generation, diffusion-based image editing techniques have gained prominence \cite{shuai2024survey}. For instance, Prompt to Prompt (P2P) \cite{hertz2022prompt} leverages the cross-attention mechanism to enable text-driven image editing, allowing precise modifications through textual prompts. To address reconstruction challenges during editing, Null-text inversion \cite{mokady2023null} achieves near-perfect reconstruction by optimizing unconditional embeddings at each denoising step. Text-driven Blended Diffusion \cite{avrahami2022blended} uses masks to target specific regions, enabling background replacement and localized edits while maintaining consistency. These advancements have also inspired adversarial attack methods \citep{xue2024diffusion,chen2024content}, which utilize image editing techniques to craft visually coherent adversarial examples that deceive detection models while preserving natural appearance.
\section{Proposed Approach}
\label{sec:appro}
In this section, we begin with an overview of our proposed attack framework, followed by a detailed introduction to its three key components.

\paragraph{Overview}
Figure \ref{fig:overview} illustrates the overall framework of BadPatch. It optimizes Null-text embeddings along incomplete diffusion trajectories and maps the image to latent vectors at intermediate timestamps, ensuring that the patch retains a natural appearance during updates. It employs \emph{Incomplete Diffusion Optimization (IDO)} guided by an IoU-Detection Loss to optimize the latent vectors, achieving faster convergence and improved attack performance. To address memory constraints, we skip the U-Net gradients during optimization and use target masks to eliminate semantic interference from irrelevant regions, enabling the generation of more naturalistic adversarial patches. Next, we will introduce the key techniques of BadPatch.

\subsection{Image-to-Latent Mapping}
Our goal is to achieve a precise mapping between a given image $\mathcal{I}$ and its corresponding latent vector $z_t$. This is accomplished by leveraging the image encoding $z_0$ to obtain $z_t$, which is essentially the reverse process of DDIM sampling \cite{song2020denoising}. Based on the assumption that the ODE process can be reversed in the limit of small steps, DDIM inversion can be formulated as:
\begin{equation}
\resizebox{\columnwidth}{!}{$ \displaystyle
    z_{t+1} = \sqrt{\frac{\alpha_{t+1}}{\alpha_t}}z_t + \left(\sqrt{\frac{1}{\alpha_{t+1}}-1}-\sqrt{\frac{1}{\alpha_t}-1}\right) \cdot \varepsilon_\theta(z_t,t,\mathcal{C}),
    $}
\label{eq:ddim_inversion}
\end{equation}
where $\varepsilon_\theta$ is the pre-trained U-Net for noise prediction, $\mathcal{C}$ is the condition embedding of $\mathcal{P}$, and $\alpha_t$ is calculated based on the schedule $\beta_0,\ldots,\beta_T\in (0,1)$, i.e., $\alpha_t=\prod_{i=1}^{t}(1-\beta_i)$. While this formula allows us to obtain the latent vector $z_t$ and reconstruct the image, it does not guarantee perfect reconstruction.

To improve the quality of generated outputs, Stable Diffusion employs a classifier-free guidance mechanism \cite{ho2022classifier} , which combines unconditional generation with guided generation based on condition embedding $\mathcal{C}$:
\begin{equation}
    \tilde{\varepsilon}_\theta(z_t,t,\mathcal{C},\phi) = w \cdot \varepsilon_\theta(z_t,t,\mathcal{C}) + (1-w) \ \cdot \varepsilon_\theta(z_t,t,\phi),
\end{equation}
where $w$ is the guidance scale parameter, and $\phi = \mathcal{E}("")$ is the embedding of a null text. The generated content is highly sensitive to textual prompts, and even minor variations can cause trajectory deviations during guided generation. These deviations accumulate, making reconstruction challenging, especially during image editing.

To address this, we separately optimize the null text embedding $\phi_t$ for each timestamp $t$ in the diffusion process $t = T \rightarrow t = 1$, guiding the process back to the correct trajectory.
Similar to Null-text inversion \cite{mokady2023null}, we first set the guidance scale $w$ to 1 and use DDIM inversion to obtain intermediate vectors $z^*_T,\ldots,z^*_0$, where $z^*_0=z_0$. Treating $z^*_T$ as the initial noise vector $\bar{z}_T$, we optimize $\phi_t$ using the following objective function after each iteration of the DDIM sampling step (with the guidance scale $w$ set to the default value of 7.5):
\begin{equation}
    \min_{\phi_t} \| z^*_{t-1}-z_{t-1}(\bar{z_t}, \phi_t, \mathcal{C}) \|^2_2,
\end{equation}
where $z_{t-1}$ is the vector obtained after DDIM sampling. After each optimization step, we update $\bar{z}_{t-1}=z_{t-1}$. This process enables near-perfect reconstruction using the initial noise vector $\bar{z}_T$ and the optimized unconditional embeddings $\{\phi_t\}^T_{t=1}$.

\subsection{Incomplete Diffusion Optimization}

We propose \emph{Incomplete Diffusion Optimization (IDO)}, a method that introduces adversarial perturbations along the trajectory of image reconstruction. IDO comprises two key techniques: 1) IoU-Detection Loss and 2) Incomplete Diffusion Process.

\paragraph{IoU-Detection Loss} 
When an intermediate adversarial image is input into the detector, it produces three outputs: 1) $B_{box}$: the coordinates of the predicted bounding box; 2) $P_{obj}$: the probability of an object being present within the predicted box; and 3) $P_{cls}$: the classification probabilities for different objects. Previous work \cite{thys2019fooling} employed the Common Detection Loss to guide adversarial patch generation:
\begin{equation}\label{eq:det_loss}
     \mathcal{L}_{det}=\frac{1}{N}\sum^{N}_{i=1}\max_{\substack{j}}(P^j_{obj}(I'_i) \cdot P^j_{cls}(I'_i)),
\end{equation}
where $N$ is the batch size, and $I'_i$ denote the $i$-th image in the batch. For each image $I'_i$, the detector identifies multiple objects, assigning an index $j$ to each detection. The images $\{I'_i\}_{i=1}^{N}$ are generated by applying an operation $\mathcal{T}$ on clean images, which involves pasting patches onto the clothing of individuals in the images.

The loss function in Eq.\eqref{eq:det_loss} optimizes each image $I'_i$ by focusing solely on the object with the highest confidence product, aiming to minimize its detection probability. However, this approach presents several limitations. First, the effectiveness of adversarial patches depends heavily on the availability and accuracy of dataset labels. Incomplete annotations or misdetections can leave critical surfaces unpatched, hindering the optimization process. Second, concentrating the loss on a single object per iteration slows convergence and increases training instability, reducing the overall effectiveness of the method.

We propose an alternative loss function \textbf{IoU-Detection Loss} to address these limitations. Formally, it is defined as:
\begin{equation}\label{eq:iou_loss}
\resizebox{\columnwidth}{!}{$ \displaystyle
     \mathcal{L}_{IoU} = \frac{1}{N} \sum_{i=1}^N \left( \frac{1}{M} \sum_{k=1}^{K} \left[ \mathbbm{1} \left(\max_{j} \mathrm{IoU}(J_j, J'_k) > t \right) P(J'_k) \right] \right),
     $}
\end{equation}
where $M$ is the total number of detected bounding boxes with an Intersection over Union (IoU) greater than a threshold $t$ with any ground truth box, $\mathrm{IoU}(J_j, J'_k)$ represents the IoU between the predicted bounding box $J'_k$ and the ground truth box $J_j$, $\mathbbm{1}(\cdot)$ is an indicator function that equals 1 if the condition inside the parentheses is met and 0 otherwise, and $P$ denotes the product of the object probability $P_{obj}$ and the classification probability $P_{cls}$. The IoU-Detection Loss accounts for multiple objects with patches applied to a single image and mitigates the impact of non-patched objects on the training process.

\paragraph{Incomplete Diffusion Process} We observed that using the default 50-step sampling to generate patches often results in a loss of semantic meaning and a diminished natural appearance during optimization. Excessive sampling steps can amplify subtle changes in $z_T$, causing significant deviations in the image reconstruction trajectory. To address this, we employ an incomplete diffusion process, where the image is matched to the latent vector $z_{\frac{T}{2}}$ at an intermediate timestamp during the image-to-latent mapping phase. This approach reduces cumulative errors, preserving the patch's semantic integrity and enhancing its natural appearance. The generation of the adversarial patch $z_p$ can then be represented as:
\begin{equation}
     z_p = G(\bar{z}_{\frac{T}{2}},\frac{T}{2},\mathcal{C},\{\phi_t\}^{\frac{T}{2}}_{t=1}),
\end{equation}
where $G$ represents the diffusion model. By generating patches using $z_{\frac{T}{2}}$, we shorten the generation path, improving the stability of the patches under adversarial guidance. This method effectively minimizes semantic loss while maintaining naturalness. However, as $z_{\frac{T}{2}}$ is updated, it becomes more chaotic, making complete denoising increasingly challenging in later stages. To prevent noticeable adversarial perturbations, we constrain the perturbations $\delta$ applied to the latent vector $z_{\frac{T}{2}}$:
\begin{equation}
     \delta_t = \text{Proj}_\infty(\delta_{t-1}+\Delta\delta,0,\epsilon),
\end{equation}
where the perturbation $\delta$ is constrained within a sphere centered at 0 with a radius of $\epsilon$ through the projection function. This ensures that the internal semantics of $z_{\frac{T}{2}}$ are not excessively lost, enabling the generation of patches with natural appearances through the denoising process.

Specifically, during denoising, 25 steps of DDIM sampling are required, involving 25 noise predictions by the U-Net. The denoising process removes noise at timestamp $t$ using a sampling recurrence formula to obtain the vector $z_{t-1}$, gradually leading $z_0$, and finally uses an image decoder $\Psi$ to restore the image. During the gradient update process, each step of DDIM sampling can be expressed as:
\begin{equation}
\resizebox{\columnwidth}{!}{$ \displaystyle
     \frac{\partial z_{t-1}}{\partial z_t} = \sqrt{\frac{\alpha_{t-1}}{\alpha_t}} + \left(\sqrt{\frac{1}{\alpha_{t-1}} - 1} - \sqrt{\frac{1}{\alpha_t} - 1}\right) \cdot \frac{\partial \varepsilon_\theta(z_t, t, \mathcal{C})}{\partial z_t}
     $}.
\end{equation}

Each denoising step requires computing and storing the U-Net gradient, which can lead to memory overflow. To address this, we adopt the approach from previous work \cite{chen2024content} by bypassing the gradient computation of the U-Net, thereby obtaining an approximate gradient. When optimizing the vector $z_{T/2}$ with the loss function $\mathcal{L}$, it can be expanded using the chain rule as follows:
\begin{equation}
     \nabla_{z_{T/2}} \mathcal{L} \approx \frac{\partial \mathcal{L}}{\partial x} \cdot \frac{\partial x}{\partial z_p} \cdot \frac{\partial z_p}{\partial z_0} \cdot \sqrt{\frac{1}{\alpha_1}} \cdot \cdots \cdot \sqrt{\frac{\alpha_{T/2-1}}{\alpha_{T/2}}}.
\end{equation}
where $z_p$ represents the adversarial patch obtained through the image encoder $\Psi$, i.e., $z_p=\Psi(z_0)$ , and $x$ is the training data with $z_p$ attached.

\subsection{Target Mask Control} Our method can generate adversarial patches of various shapes, controlled by the target mask $m$. Using target masks not only allows for the extraction of patches of different shapes but also plays a crucial role in maintaining the natural appearance of the patch. Since Stable Diffusion \cite{rombach2022high} generates square images, there is environmental information surrounding our target. After selecting the initial image $\mathcal{I}$, to prevent the background from interfering with the main subject during the optimization process, we replace the background with a solid-colored image $s$ before the Image-to-Latent Mapping stage:
\begin{equation}
     \mathcal{I}' = \mathcal{I} \odot m + s \odot (1-m).
\end{equation}
Additionally, to ensure that the background remains unchanged throughout the update process, we downsample the mask $m$ by a factor of 8 to match the size of latent vectors and overlay gradients outside the target area.

\begin{figure}[!ht]
    \centering
    \includegraphics[width=\columnwidth]{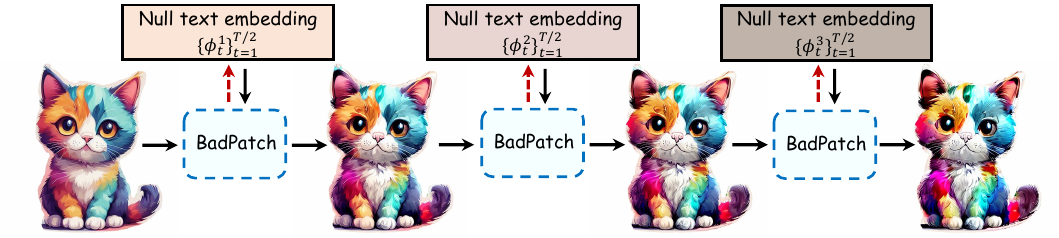}
    \caption{The effect of iterative generation (showing accumulated style variation from left to right).}
    \label{fig:nested_opt}
    \vspace{-0.4cm}
\end{figure}

Overall, the BadPatch generation process begins with Image-to-Latent Mapping, followed by Incomplete Diffusion Optimization to reduce memory consumption and enhance performance. Finally, Target Mask Control is applied to preserve the natural appearance of the generated patches. The complete generation process is outlined in Algorithm \ref{alg:diffpatch}. Furthermore, the patch can be iteratively optimized using BadPatch to achieve an accumulative effect, as illustrated in Figure \ref{fig:nested_opt}. Specifically, the generated patch can serve as the input for generating a new patch, enabling stronger attack performance while maintaining a natural appearance. This iterative approach ensures both effectiveness and stealthiness in adversarial patch generation.

\begin{algorithm}[tb]
\caption{BadPatch Generation Process}
\label{alg:diffpatch}
\textbf{Input}: A patch image $\mathcal{I}'$ with target mask $m$, prompt embedding $\mathcal{C}=\mathcal{E}(\mathcal{P})$, an object detector $\mathcal{D}$, training dataset $S$ and ground truth box $y$.\\
\textbf{Parameter}: DDIM steps $T$, unconditional embeddings iteration number $N_u$, training batch $B_t$, and patch scale $\tau$.\\
\textbf{Output}: Adversarial patch
\begin{algorithmic}[1] 
\STATE Compute $z^*_{\frac{T}{2}},\ldots,z^*_0$ using DDIM inversion over $\mathcal{I}'$ with guidance scale $w=1$
\STATE Set $w=7.5$, $\bar{z}_{\frac{T}{2}} \leftarrow z^*_{\frac{T}{2}}$, $\phi_{\frac{T}{2}} \leftarrow \mathcal{E}(""),\delta_0 \leftarrow 0$
\FOR{$t=\frac{T}{2},\frac{T}{2}-1,\ldots,1$}
\FOR{$j=0,\ldots,N_u-1$}
\STATE $\phi_t \leftarrow  \phi_t - \eta_u \nabla_{\phi} \|{z^*_{t-1}-z_{t-1}(\bar{z_t}, \phi_t, \mathcal{C})}\|_2^2$
\ENDFOR
\STATE $\bar{z}_{t-1} \leftarrow z_{t-1}(\bar{z_t}, \phi_t, \mathcal{C})$, $\phi_{t-1} \leftarrow  \phi_{t} $
\ENDFOR
\STATE $m^* \leftarrow Downsample(m)$
\FOR{$k=1,\ldots,N_{iter}$}
\FOR{$b=1,\ldots,B_t$}
\STATE $X \leftarrow$ Extract the $b$-th batch of data from $S$
\STATE $z_p \leftarrow G(\bar{z}_{\frac{T}{2}}+\delta_{b-1},\frac{T}{2},\mathcal{C},\{\phi_t\}^{\frac{T}{2}}_{t=1})\odot m$
\STATE $X' \leftarrow \mathcal{T}(z_p, X, \tau)$
\STATE $g_b \leftarrow Adam(\nabla_{\bar{z}_{\frac{T}{2}}} \mathcal{L}_{IoU}(\mathcal{D}(X'),y))$
\STATE $\delta_b \leftarrow Proj_\infty(\delta_{b-1}+\eta_p \cdot g_b, 0,\epsilon) \odot m^*$
\ENDFOR
\STATE Set $\delta_0=\delta_B$
\ENDFOR
\STATE $z_p \leftarrow G(\bar{z}_{\frac{T}{2}}+\delta_B,\frac{T}{2},\mathcal{C},\{\phi_t\}^{\frac{T}{2}}_{t=1})\odot m$
\STATE \textbf{return} $z_p$
\end{algorithmic}
\end{algorithm}
\section{Experiments}
\label{sec:exp}

\begin{table*}[ht]
  \centering

  \caption{Comparing BadPatch with other methods in ASR (\%) (higher is better). The best results are \underline{\textbf{underlined}}. $'$ denotes transfer attack methods, while the rest are white-box attacks. $^\dag$ trained on YLv2, $^*$ trained on YLv4t. The generated patches are shown at the bottom.}
\small
  \begin{tabular*}{\textwidth}{l@{\extracolsep{\fill}}ccccccccccc}
    \toprule
    \multirow{2}{*}{Method} & \multicolumn{11}{c}{ASR (\% $\uparrow$)} \\
    \cmidrule(lr){2-12}
     & YLv3 & YLv3t & YLv4 & YLv4t & YLv5s & YLv7t &YLv10s &YLv12s & FRCNN & DETR & RT-DETR \\
    \midrule
    $^{(P_A)}$AdvYL$^\dag$ & 46.3 & 66.4 & 36.2 & 69.0 & 53.4 & 20.7 & 35.1 & 30.6 & 24.0 & 36.7 & 30.6 \\
    $^{(P_B)}$T-SEA$'$$^*$  & 39.0 & 59.0 & 48.7 & 56.8 & 49.3 & 26.6 & 61.1   & 62.1 & 27.1 & 36.3  & 32.1\\
    $^{(P_C)}$NPAP$^*$    & 33.7 & 74.2 & 26.2 & 78.9 & 41.4 & 17.5 & 33.4  & 22.0 & 25.1 & 41.5  & 48.3\\
    $^{(P_D)}$D2D$^*$    & 33.9 & 69.1 & 30.9 & 75.0 & 40.7 & 18.4 & 23.1 &26.0 &  25.0  &34.2  & 39.7\\
    $^{(P_E)}$DAP$^*$  & 39.0 & 49.5 & 32.9 & 46.0 & 43 & 18.3 & 20.7 & 23.0 & 22.9 & 39.8 & 25.6\\
    $^{(P_F)}$AdvART$^*$  & 22.9 & 36.4 & 15.4 & 33.9 & 19.3 & 17.3 & 12.1 & 11.4 & 20.4 & 31.0 & 19.0\\
    \midrule
    $^{(P_1)}$YLv3(Ours)  & \underline{\textbf{70.4}} & 64.1 & 28.1 & 60.4 & 52.2 & 24.6 & 26.6  & 26.8& 29.3  &38.6  & 44.9\\
    $^{(P_2)}$YLv3t(Ours) & 38.6 & \underline{\textbf{86.1}} & 24.4 & 65.3 & 44.8 & 24.1 & 24.0 & 19.9 & 29.5  &39.6  & 46.7\\
    $^{(P_3)}$YLv4(Ours)  & 45.0 & 57.5 & \underline{\textbf{49.9}} & 46.8 & 43.5 & 21.6 & 27.5 & 24.8 & 27.4  &39.0  & 45.2\\
    $^{(P_4)}$YLv4t(Ours) & 41.0 & 73.6 & 25.5 & \underline{\textbf{79.1}} & 50.3 & 25.7 & 27.8  &21.0 & 28.4  &41.4  & 46.5\\
    $^{(P_5)}$YLv5s(Ours) & 47.4 & 64.3 & 27.2 & 56.5 & \underline{\textbf{69.9}} & 23.1 & 31.6  & 25.7 & 29.9  &38.3  & 37.7\\
    $^{(P_6)}$YLv7t(Ours) & 35.0 & 64.3 & 23.0 & 56.7 & 44.5 & \underline{\textbf{33.5}} & 31.6  &25.3 & 26.1  &38.9  & 47.7\\
    $^{(P_7)}$YLv10s(Ours)  & 35.3 & 62.0 & 23.4 & 59.9 & 45.7 & 24.2 & \underline{\textbf{66.1}} & 27.2 & 28.9 & 40.3 &  45.0\\
    $^{(P_8)}$YLv12s(Ours)  & 44.0 & 57.3 & 37.9 & 55.1 & 45.1 & 22.2 & 47.7 & \underline{\textbf{63.7}} & 32.4 & 27.4 & 42.7\\
    $^{(P_9)}$FRCNN(Ours)& 42.7 & 62.4 & 29.2 & 60.8 & 50.7 & 24.3 & 28.8  & 24.8 & \underline{\textbf{38.5}}  &43.7  & 45.5\\
    $^{(P_{10})}$DETR(Ours) & 64.2 & 79.5 & 43.1 & 75.2 & 62.7 & 27.1 & 41.2  &31.8 & 37.0  &\underline{\textbf{46.5}}  & 52.0\\
    $^{(P_{11})}$RT-DETR(Ours)  & 36.9 & 55.2 & 22.2 & 48.5 & 40.1 & 22.3 & 26.7 & 23.5 & 27.1 & 37.9 & \underline{\textbf{56.3}}\\
    \midrule
    \midrule
    $^{(P_j)}$Source    & 29.7 & 38.1 & 18.5 & 36.9 & 35.0 & 18.4 & 16.0 & 17.4&  23.3  &33.6  &27.3 \\
    \hspace{0.5cm}Gray    & 16.5 & 25.2 & 12.8 & 30.0 & 15.5 & 13.5 & 13.2 & 12.3&  17.3  &34.4  & 16.5\\
    $^{(P_m)}$Random    & 16.9 & 22.2 & 12.2 & 26.9 & 15.5 & 12.3 & 11.2 & 12.7&  17.6  &31.7  & 15.6\\
    $^{(P_n)}$Random$'$ & 17.0 & 19.0 & 10.4 & 20.7 & 11.7 & 12.0 & 9.8 &10.6 &  16.4  &25.9  & 15.5\\
    \bottomrule
  \end{tabular*}
    
    \includegraphics[width=0.07\textwidth]{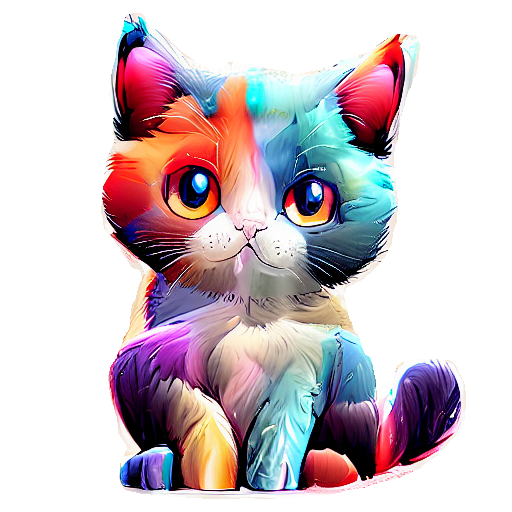} 
    \hspace{0.02\textwidth}
    \includegraphics[width=0.07\textwidth]{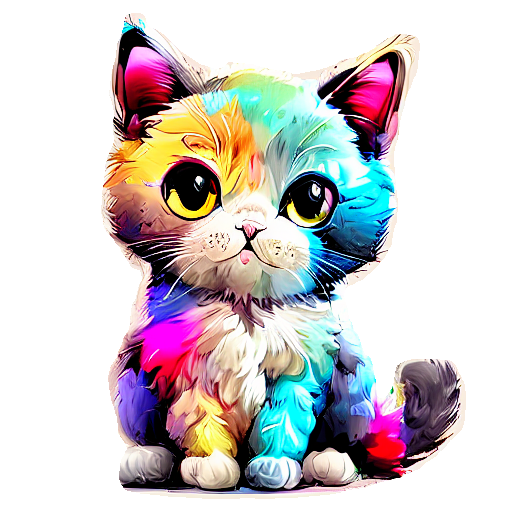}
    \hspace{0.02\textwidth}
    \includegraphics[width=0.07\textwidth]{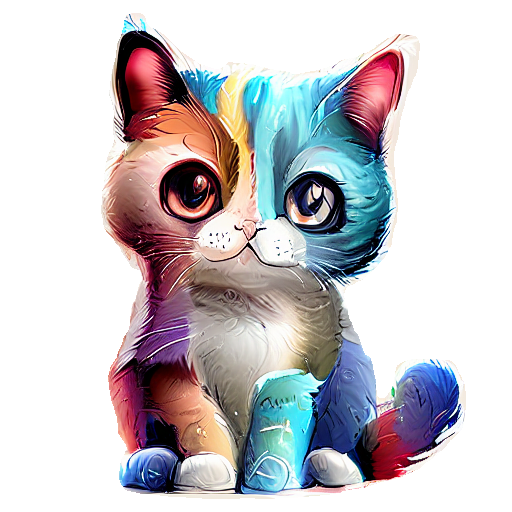}
    \hspace{0.02\textwidth}
    \includegraphics[width=0.07\textwidth]{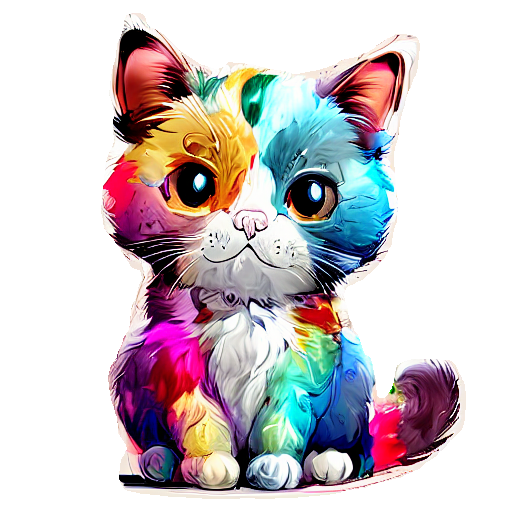}
    \hspace{0.02\textwidth}
    \includegraphics[width=0.07\textwidth]{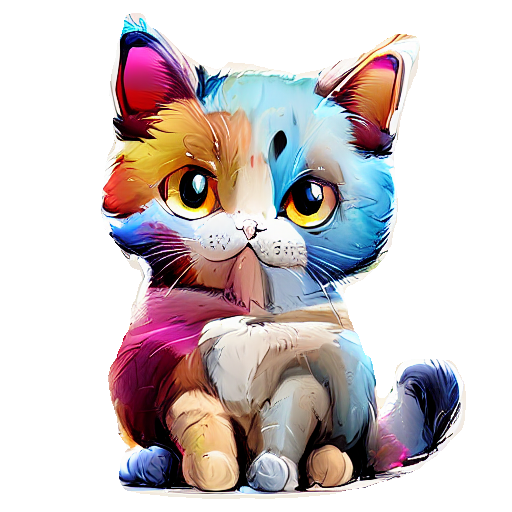}
    \hspace{0.02\textwidth}
    \includegraphics[width=0.07\textwidth]{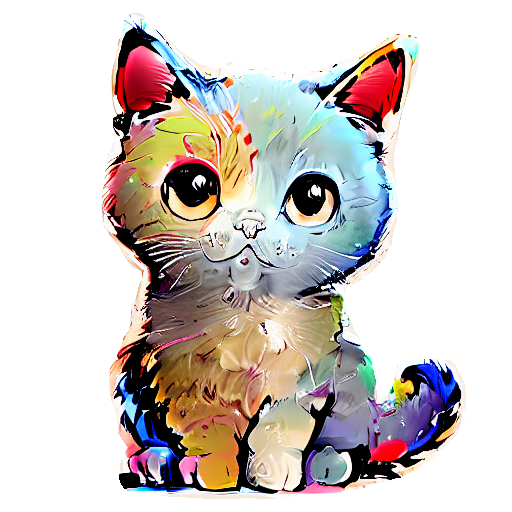}
    \hspace{0.02\textwidth}
    \includegraphics[width=0.07\textwidth]{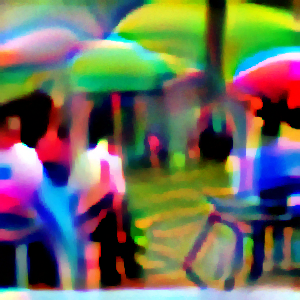}
    \hspace{0.02\textwidth}
    \includegraphics[width=0.07\textwidth]{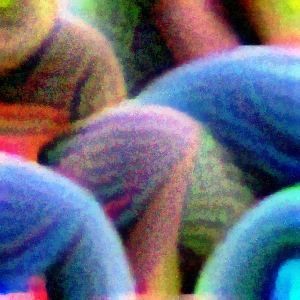}
    \hspace{0.02\textwidth}
    \includegraphics[width=0.07\textwidth]{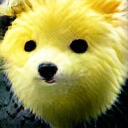}
    \hspace{0.02\textwidth}
    \includegraphics[width=0.07\textwidth]{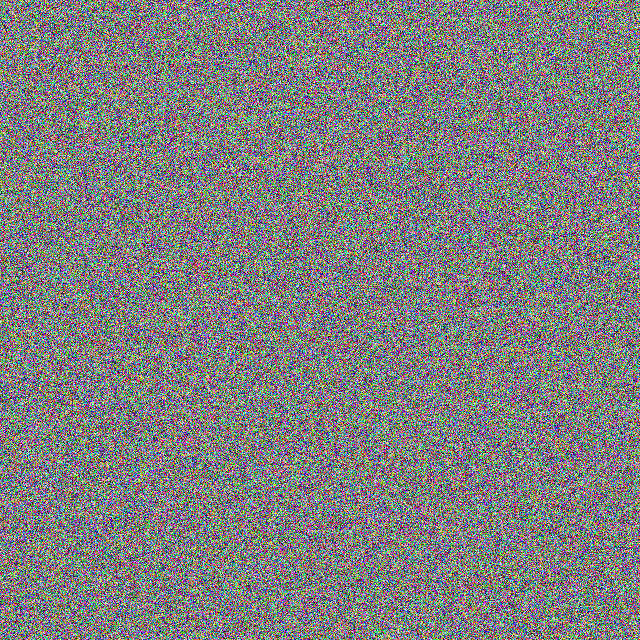}\\
    
    \makebox[0.07\textwidth][c]{\scriptsize $P_1$}
    \hspace{0.02\textwidth}
    \makebox[0.07\textwidth][c]{\scriptsize $P_2$}
    \hspace{0.02\textwidth}
    \makebox[0.07\textwidth][c]{\scriptsize $P_3$}
    \hspace{0.02\textwidth}
    \makebox[0.07\textwidth][c]{\scriptsize $P_4$}
    \hspace{0.02\textwidth}
    \makebox[0.07\textwidth][c]{\scriptsize $P_5$}
    \hspace{0.02\textwidth}
    \makebox[0.07\textwidth][c]{\scriptsize $P_6$}
    \hspace{0.02\textwidth}
    \makebox[0.07\textwidth][c]{\scriptsize $P_A$}
    \hspace{0.02\textwidth}
    \makebox[0.07\textwidth][c]{\scriptsize $P_B$}
    \hspace{0.02\textwidth}
    \makebox[0.07\textwidth][c]{\scriptsize $P_C$}
    \hspace{0.02\textwidth}
    \makebox[0.07\textwidth][c]{\scriptsize $P_m$}\\
    \includegraphics[width=0.07\textwidth]{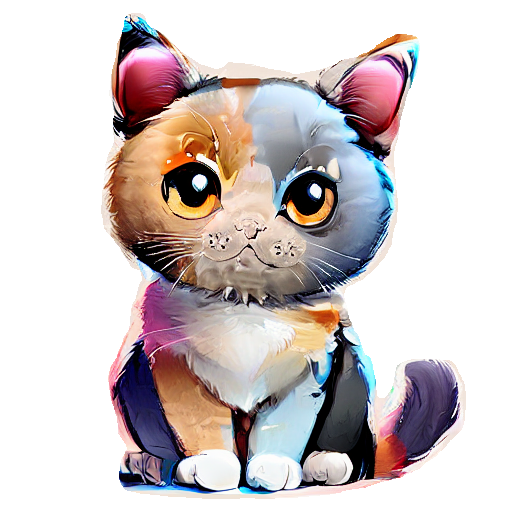}
    \hspace{0.02\textwidth}
    \includegraphics[width=0.07\textwidth]{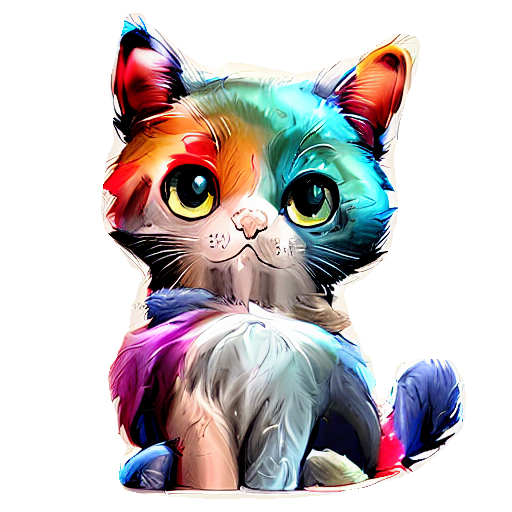}
    \hspace{0.02\textwidth}
    \includegraphics[width=0.07\textwidth]{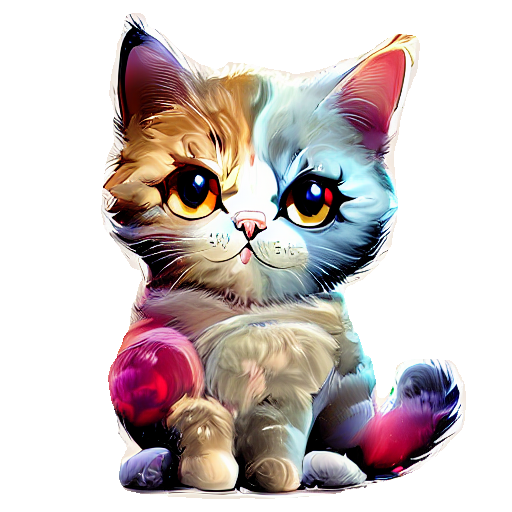}
    \hspace{0.02\textwidth}
    \includegraphics[width=0.07\textwidth]{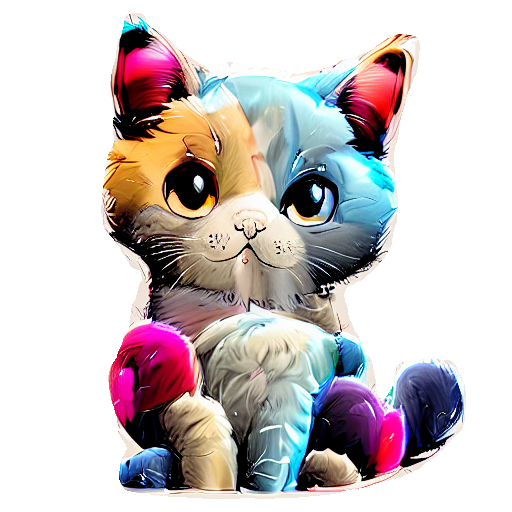}
    \hspace{0.02\textwidth}
    \includegraphics[width=0.07\textwidth]{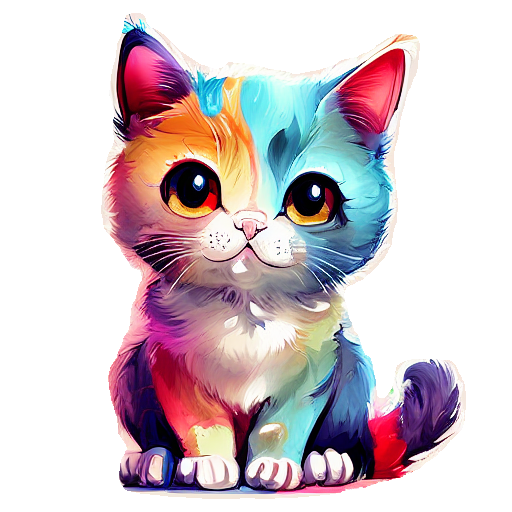}
    \hspace{0.02\textwidth}
    \includegraphics[width=0.07\textwidth]{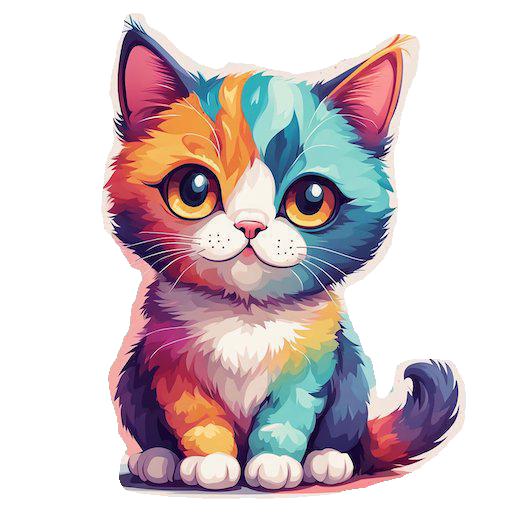}
    \hspace{0.02\textwidth}
    \includegraphics[width=0.07\textwidth]{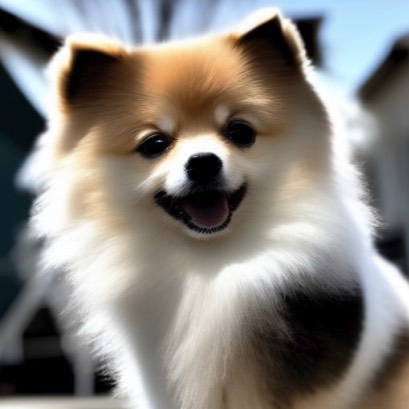}    \hspace{0.02\textwidth}
    \includegraphics[width=0.07\textwidth]{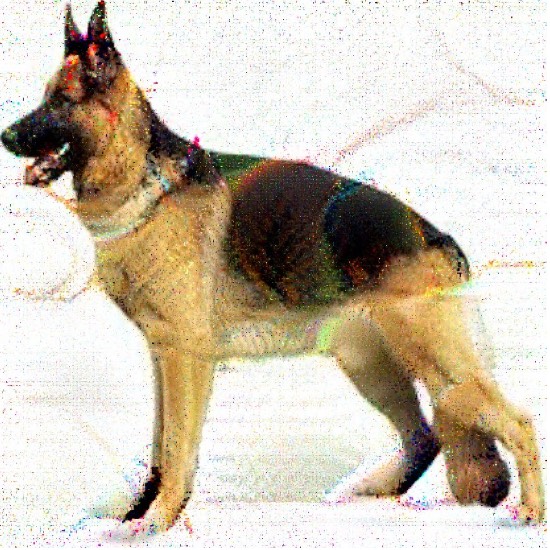}
    \hspace{0.02\textwidth}
    \includegraphics[width=0.07\textwidth]{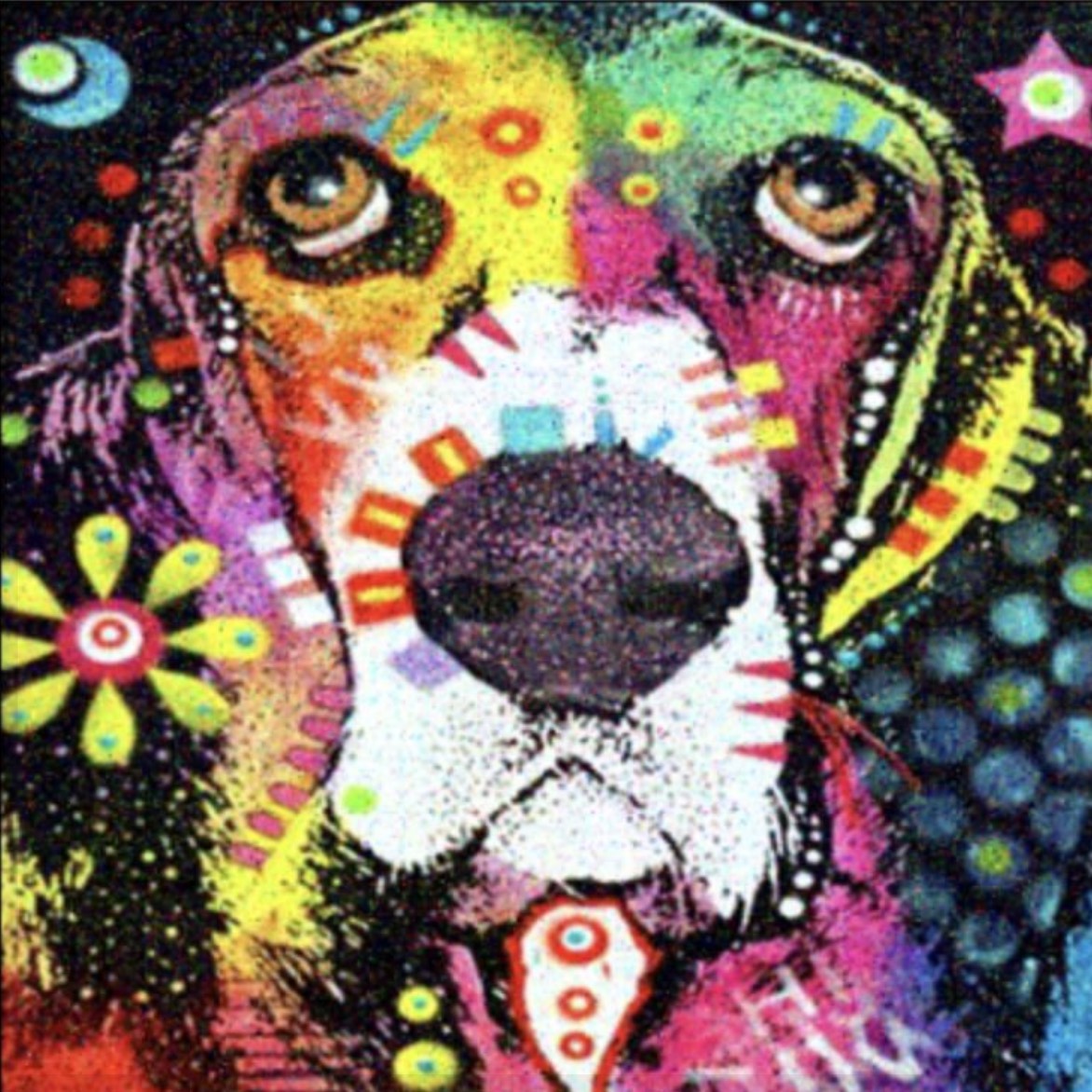}    \hspace{0.02\textwidth}
    \includegraphics[width=0.07\textwidth]{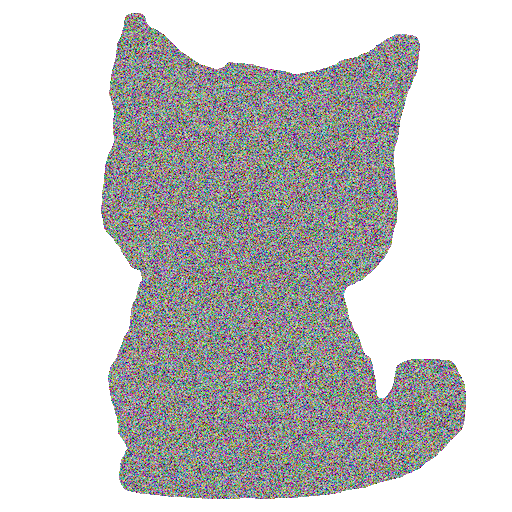}\\
    \makebox[0.07\textwidth][c]{\scriptsize $P_7$}
    \hspace{0.02\textwidth}
    \makebox[0.07\textwidth][c]{\scriptsize $P_8$}
    \hspace{0.02\textwidth}
    \makebox[0.07\textwidth][c]{\scriptsize $P_9$}
    \hspace{0.02\textwidth}
    \makebox[0.07\textwidth][c]{\scriptsize $P_{10}$}
    \hspace{0.02\textwidth}
    \makebox[0.07\textwidth][c]{\scriptsize $P_{11}$}
    \hspace{0.02\textwidth}
    \makebox[0.07\textwidth][c]{\scriptsize $P_j$}
    \hspace{0.02\textwidth}
    \makebox[0.07\textwidth][c]{\scriptsize $P_D$}
    \hspace{0.02\textwidth}
    \makebox[0.07\textwidth][c]{\scriptsize $P_E$}
    \hspace{0.02\textwidth}
    \makebox[0.07\textwidth][c]{\scriptsize $P_F$}
    \hspace{0.02\textwidth}
    \makebox[0.07\textwidth][c]{\scriptsize $P_n$}
  \label{tab:eva}%
  \vspace{-0.2cm}
\end{table*}%

\subsection{Experimental Setup}
\paragraph{Datasets} In our experiments, we train and evaluate the adversarial patches on the INRIA person dataset \cite{dalal2005histograms}, which consists of 614 training images and 288 test images. To meet the input requirements of the detector, we pad all images with gray pixels to make them square and then resize them to 640 $\times$ 640. 
Furthermore, we evaluate the transferability of different adversarial patch methods on the MPII Human Pose dataset \cite{andriluka20142d}, COCO dataset \cite{lin2014microsoft} and 1,000 images from diverse scenarios collected from the Internet.

\paragraph{Victim Models} Our target a variety of detectors, including YOLOv3, YOLOv3-tiny \cite{redmon2018yolov3}, YOLOv4, YOLOv4-tiny \cite{bochkovskiy2020yolov4}, YOLOv5s \cite{glenn_jocher_2020_4154370}, YOLOv7-tiny \cite{wang2023yolov7}, YOLOv10s \cite{wang2025yolov10}, YOLOv12s \cite{tian2025yolov12}, FasterRCNN \cite{ren2015faster} and DETR \cite{carion2020end} and RT-DETR \cite{zhao2024detrs}. We will abbreviate YOLO as YL and FasterRCNN as FRCNN for ease of presentation, such as YOLOv4-tiny as YLv4t. All models were pre-trained on the COCO dataset \cite{lin2014microsoft}.

\paragraph{Implementation Details}
We treat the person as the target object class, setting the batch size to 32, the maximum number of epochs $N_{iter}=200$, DDIM sampling steps $T=50$, unconditional embedding optimization iterations $N_u=10$, learning rates $\eta_u=0.01$ and $\eta_p=0.003$, and perturbation constraint $\epsilon=0.5$. We use the Adam optimizer \cite{kingma2014adam} with its default parameters. Textual prompts $\mathcal{P}$ are automatically generated using BLIP \cite{li2022blip}. The adopted version of Stable Diffusion \cite{rombach2022high} is v1.4. Since we generate irregular patches, we set the patch size to match the pixel area of a square patch with a scale of $\tau=0.2$. To achieve stronger performance, we apply two iterations of optimization to the patch using BadPatch. 
We evaluate the effectiveness of different patch generation methods using both the Attack Success Rate (ASR) and Average Precision (AP). The ASR measures the success of the patches in misleading the detector, while AP reflects the overall decline in detection accuracy. The confidence threshold and IoU threshold for the detector are both set to 0.5. All experiments were conducted on a single NVIDIA A100 GPU.

\subsection{Attack Performance Evaluation}
We generate a series of adversarial patches targeting different models using BadPatch and compare them on the INRIA dataset with existing methods, including AdvYL \cite{thys2019fooling}, T-SEA \cite{huang2023t}, DAP \cite{guesmi2024dap}, AdvART \cite{guesmi2024advart}, NPAP \cite{hu2021naturalistic}, and D2D \cite{lin2023diffusion}. The detailed results are reported in Table \ref{tab:eva}. As shown, BadPatch ($P_1-P_{11}$) achieves the highest Attack Success Rate (ASR), improving by \textbf{10.1\%} over AdvYL on YOLOv4-tiny. When compared to naturalistic adversarial patches ($P_C$ and $P_D$), our patch trained on YOLOv4 results in a \textbf{0.2\%} and \textbf{4.1\%} ASR increase, respectively. Additionally, as demonstrated in the Appendix B, BadPatch exhibits strong performance in the AP metric, performing comparably to unnatural patch generation methods.

\subsection{Cross-dataset Evaluation}
Given the relatively simple scenes in the INRIA dataset, we further evaluate the attack performance of the patches in more diverse environments. We train patches using YOLOv4 with various methods and test them across multiple datasets, including MPII and COCO. Due to the large size of these datasets, we randomly sample a subset of images containing people for testing, specifically 729 images from COCO and 569 images from MPII. 
Additionally, since the images from these two datasets are relatively outdated and of low resolution, we also collect 1,000 high-resolution images from the Internet, representing diverse scenes (denoted as "Collected"). As shown in Table \ref{tab:cross_data}, our method significantly reduces the AP while substantially increasing the ASR. On average, the ASR shows a $\textbf{17.5\%}$ improvement over the unnatural patch method T-SEA and a $\textbf{2.6\%}$ improvement over the natural patch method NPAP. Several example images and their detection results with adversarial patches are provided in the Appendix A.

\begin{table}[h]
    \centering
    \small
    \begin{tabular*}{\columnwidth}{l@{\extracolsep{\fill}}ccccc}
        \toprule
        \multirow{2}{*}{Metric} & \multirow{2}{*}{Dataset} & \multicolumn{4}{c}{Attack} \\ 
        \cmidrule(lr){3-6}
         & & T-SEA & NPAP & D2D & \textbf{Ours} \\
        \midrule
        \multirow{4}{*}{ASR\%$\uparrow$} &
        INRIA & 56.8 & 78.9 & 75.0 & \textbf{79.1} \\
        & MPII  & 60.3 & 73.4 & 73.3 & \textbf{75.9}\\
        & COCO   & 54.5 & 63.9 & 60.3 & \textbf{67.8}\\
        & Collected   & 67.2 & 82.1 & 79.4 & \textbf{85.8}\\
        \midrule
        \multirow{4}{*}{AP\%$\downarrow$} &
        INRIA & 12.2 & 13.7 & 15.6 & \textbf{9.2} \\
        & MPII  & 17.0 & 17.3 & 17.9 & \textbf{12.5} \\
        & COCO   & 21.4 & 25.1 & 26.8 & \textbf{16.9}\\
        & Collected   & 8 & 6.7 & 7.2 & \textbf{3.9}\\
        \bottomrule
    \end{tabular*}
    \caption{Attack performance of different methods across datasets.}
    \label{tab:cross_data}
    \vspace{-0.4cm}
\end{table}

\paragraph{Ensemble Attack}
We also evaluate an ensemble version of BadPatch, which trains the patch using averaged gradients from multiple models. We compare this with two other methods, NPAP and D2D, where NPAP was trained on YOLOv3 and YOLOv4, and D2D was trained on all considered models. Our ensemble BadPatch was similarly trained on all models. The results, presented in the Appendix C, demonstrate that our method outperforms NPAP and D2D by a significant margin.

\begin{figure}[!ht]
    \begin{tikzpicture}[scale=1]
        \foreach \t/\i/\n in {-/0/0,0.1/1/1,0.3/2/3,0.5/3/5,0.7/4/7,0.9/5/9}{
            \node at (\i,3.8) {\small \t};
            \foreach \u/\j in {4/3,3/2,2/1,1/0}{
                \node at (\i,\j) {\includegraphics[width=0.123\linewidth]{images/eps/\u/\u_\n.png}};
            }
        }
        \foreach \u/\j in {$P_{a1}$/3,$P_{a2}$/2,$P_{a3}$/1,$P_{a4}$/0} {
            \node at (-0.8,\j) {\small \u};
        }
        \node at (2.7,4.3) {\small Constraint \ \ $\epsilon$};
        \node at (-1.3,1.5) {\rotatebox{90}{\small $Patch$}};
    \end{tikzpicture}
    \vskip -0.3cm
    
    \begin{tikzpicture}
        \pgfplotsset{
            every tick label/.append style={font=\small},
            yticklabel={\pgfmathprintnumber[assume math mode=true]{\tick}}
        }
        \begin{axis}[
            xmin=0.5, xmax=6.5,
            ymin=20, ymax=80,
            xtick pos=top,
            axis y line*=left,
            xticklabels={,,},
            width=0.92\linewidth,
            height=4cm,
            ylabel={\small Average ASR(\%)},
            y label style={at={(axis description cs:0.12,.5)},anchor=south}
        ]
            \addplot[smooth, mark=triangle, blue] table [x=eps, y=asr, col sep=comma] {images/eps/stats.csv};
            \label{plot:ASR}
        \end{axis}
        
        \begin{axis}[
            xmin=0.5, xmax=6.5,
            ymin=35, ymax=95,
            xtick pos=top,
            axis y line*=right,
            xticklabels={,,},
            width=0.92\linewidth,
            height=4cm,
            ylabel={\small Average CLIP Sim.(\%)},
            legend columns=-1,
            y label style={at={(axis description cs:1.375,.5)},anchor=south},
            legend style={at={(0.9,0.3)}}
        ]
            \addlegendimage{/pgfplots/refstyle=plot:ASR}\addlegendentry{\footnotesize ASR}
            \addplot[smooth, mark=square, red] table [x=eps, y=clip_sim, col sep=comma] {images/eps/stats.csv};
            \addlegendentry{\footnotesize CLIP Sim.}
        \end{axis}
    \end{tikzpicture}
    \caption{The effect of optimization constraint $\epsilon$ on latent vectors.}
    \label{fig:constraint}
    \vspace{-0.4cm}
\end{figure}

\subsection{Ablation Studies}

\paragraph{Optimization Constraint}  
To ensure the natural appearance of the generated patches, we constrain each optimization step using the $L_{\infty}$ norm. The choice of $\epsilon$ will influence visual naturalness of the generated patch and its attack performance. 
To quantify the natural appearance of the patches, we use GPT-4o \cite{achiam2023gpt} to generate descriptive text for reference images (denoted by '-') and compute the CLIP similarity \cite{radford2021learning} between these descriptions and the generated patches. The corresponding descriptions can be found in the Appendix D. As illustrated in Figure \ref{fig:constraint}, we obtain patches with varying styles under different constraints using BadPatch and evaluate both the ASR and CLIP similarity for each. Based on the results, we select $\epsilon=0.5$ as the optimal value, as it achieves a favorable balance between attack performance and natural appearance.

\begin{figure}[!ht]
    \centering
    \includegraphics[width=1\columnwidth]{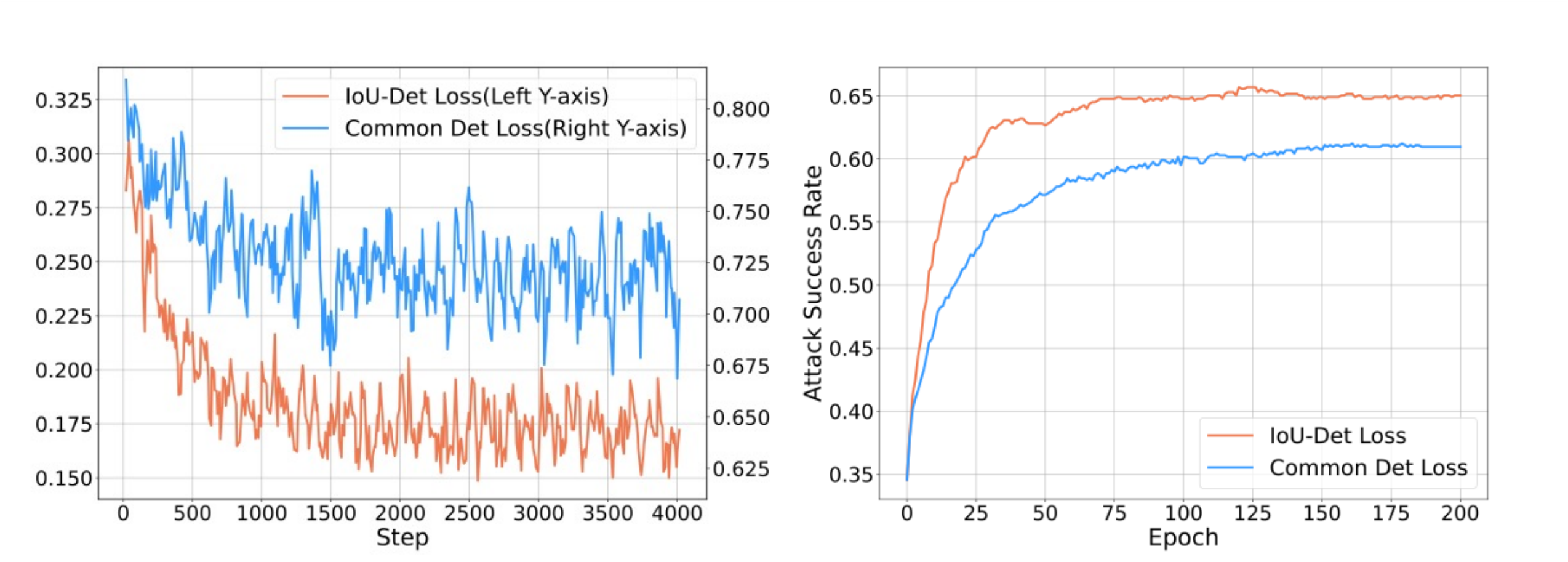}
    \caption{Comparison of two loss functions. \emph{Left}: training loss; \emph{Right}: attack performance. }
    \label{fig:loss_comp}
    \vspace{-0.4cm}
\end{figure}

\paragraph{Loss Function} Here, we compare our IoU-Detection Loss with the Common Detection Loss under the same patch training setting. As shown in Figure \ref{fig:loss_comp}, compared to the Common Detection Loss, our IoU-Detection Loss demonstrates greater stability during training process and achieves higher attack performance in an earlier time.

\paragraph{Shape Impact}  
Here, we investigate the impact of patch shapes, as used for $P_m$ and $P_n$ in Table \ref{tab:eva}. We compare the performance of random noise images with square and irregular shapes (of the same size as BadPatch). From the ASR and AP results in Table \ref{tab:eva}, it is evident that square patches yield higher attack effectiveness compared to irregular patches. In other words, designing irregular patches with high attack success rates is more challenging.

\begin{figure}[ht]
    \centering
    \includegraphics[width=0.95\columnwidth]{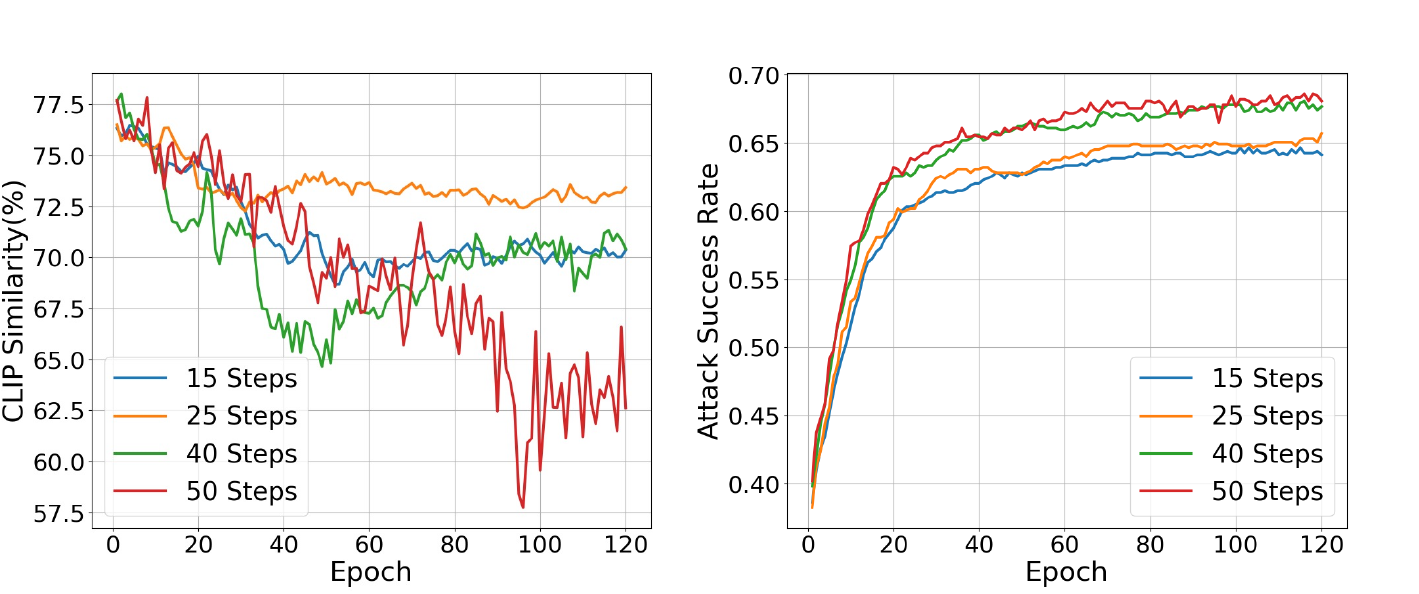}
    \caption{The effects of diffusion trajectories with different lengths on the performance of patch attacks and natural appearance.}
    \label{fig:clip_asr}
\end{figure}

\paragraph{Diffusion Trajectory Length}  
Here, we evaluate the results of latent vectors at different timestamps, assessing both the attack performance and the similarity to textual descriptions for patches generated from diffusion trajectories of varying lengths. As shown in Figure \ref{fig:clip_asr}, starting diffusion optimization from $\frac{T}{2}$ (25 steps) not only achieves a high success rate but also produces images that align more closely with textual descriptions, thereby maintaining greater semantic consistency with the original images.

\begin{figure}[ht]
    \centering
    \includegraphics[width=\columnwidth]{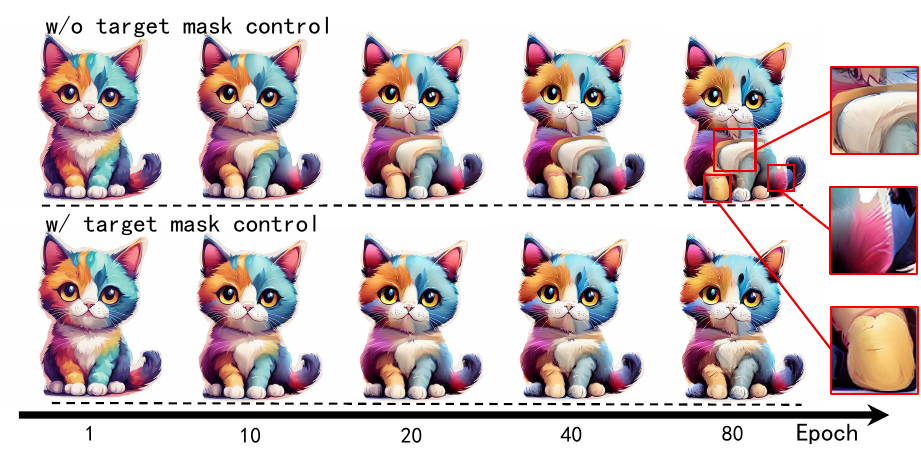}
    \caption{The impact of target mask on adversarial patch optimization. Without target mask control, adversarial patches exhibit noticeable detail distortions (indicated by the red boxes).}
    \label{fig:mask_control}
\end{figure}

\paragraph{Target Mask}  
The target mask mechanism helps mitigate semantic loss in images to some extent. Here, we compare the training results with and without the use of this mechanism. Without the target mask, the patches exhibit quality issues during optimization, as illustrated in Figure \ref{fig:mask_control}.

\paragraph{Iterative Optimization}  

As shown in Table \ref{tab:nested}, iterative optimization can further enhance the adversarial strength of the patch. However, it also tends to reduce the natural appearance of the patch.
\begin{table}[ht]
    \centering
    \small
    \begin{tabular*}{\columnwidth}{l@{\extracolsep{\fill}}ccccc}
        \toprule
        \multirow{2}{*}{Metric} & \multirow{2}{*}{Iterative} & \multicolumn{4}{c}{Model} \\ 
        \cmidrule(lr){3-6}
         & & YLv3t & YLv4t & YLv5s & FRCNN \\
        \midrule
        \multirow{2}{*}{ASR\%$\uparrow$} &
        w/o & 82.9 & 75.5 & 65.6 & \textbf{37.3} \\
        & w/  & \textbf{86.1} & \textbf{79.1} & \textbf{69.9} & \textbf{38.5}\\
        \midrule
        \multirow{2}{*}{AP\%$\downarrow$} &
        w/o & 7.1 & 12.3 & 24.6 & 26.0 \\
        & w/  & \textbf{3.9} & \textbf{9.2} & \textbf{22.0} & \textbf{22.6} \\
        \midrule
        \multirow{2}{*}{Sim.\%$\uparrow$} &
        w/o & \textbf{77.1} & \textbf{77.6} & \textbf{73.2} & \textbf{72.9} \\
        & w/  & 72.1 & 69.4 & 60.0 & 60.8 \\
        \bottomrule
    \end{tabular*}
    \caption{The impact of using iterative optimization.}
    \label{tab:nested}
\end{table}

\subsection{Adversarial T-shirt Dataset}
To better evaluate the effectiveness of our attack and create a valuable physical-world adversarial patch dataset for future research, we designed 9 physical-world adversarial T-shirts using our generated patches. We printed these patches on T-shirts and captured photos in a variety of indoor and outdoor scenarios, such as laboratory, campus, cafeteria, subway station, and shopping mall, with the assistance of 20 participants (with ethics approval). 

\begin{figure}[ht]
    \centering
    \includegraphics[width=0.9\columnwidth]{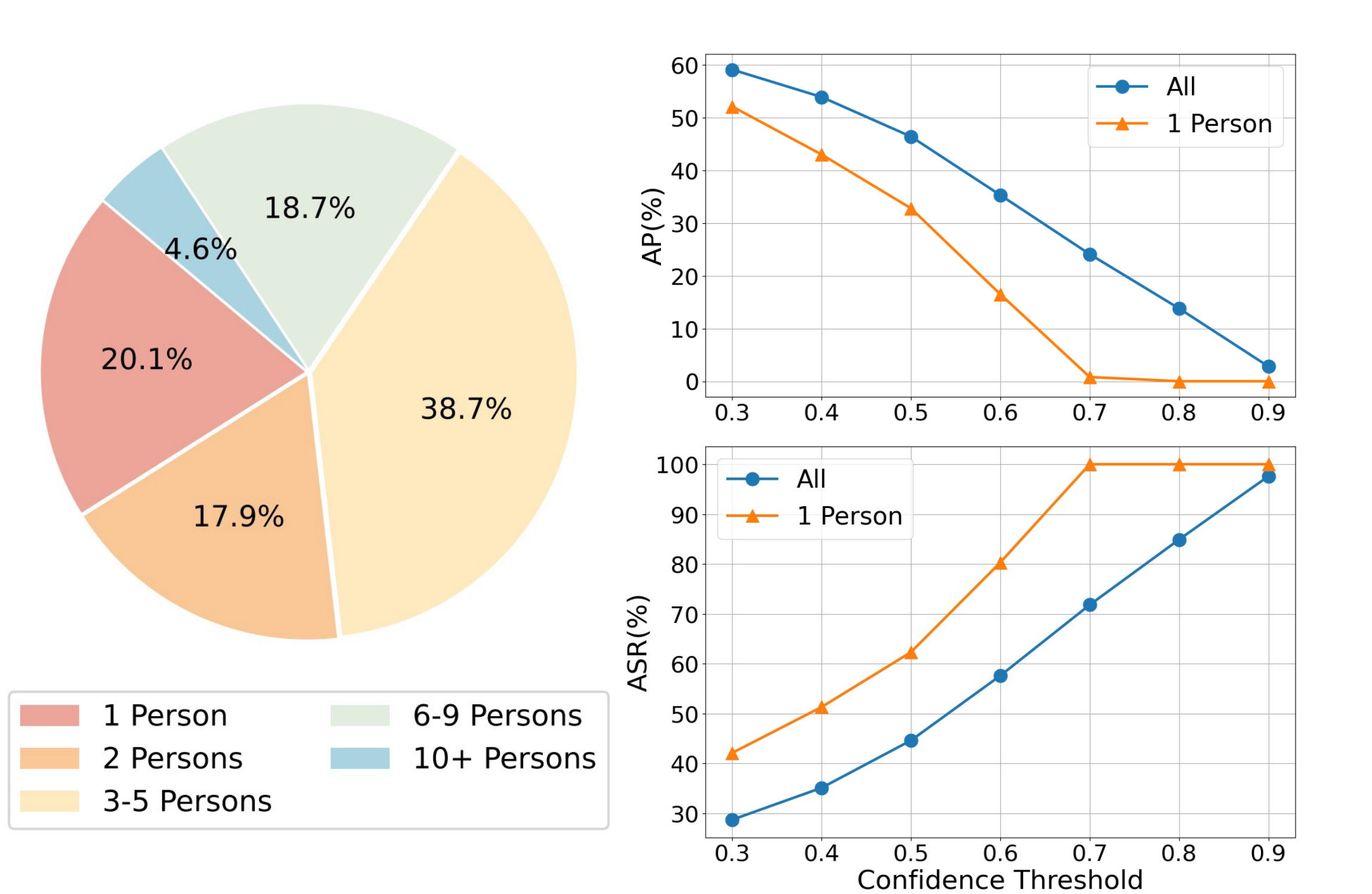}
    \caption{\emph{Left}: The distribution of the images in AdvT-shirt-1K. \emph{Right}: The detection performance of YOLOv5s on AdvT-shirt-1K under different confidence thresholds.}
    \label{fig:dataset}
\end{figure}

In total, we collected 1,131 images, including both individual and group photos (ranging from 2 to 10+ persons). We name the dataset \textbf{AdvT-shirt-1K}, which includes detailed annotations for person and patch locations (bounding boxes). 
The collection and annotation process of the dataset spanned six months, which involves the refinement of our BadPatch method for multiple times.
The statistics of the dataset is shown in Figure \ref{fig:dataset} with several example images are provided in Figure \ref{fig:subset}. The results shown in the right subfigure of Figure \ref{fig:dataset} demonstrate the effectiveness of the adversarial T-shirts in evading the YOLOv5s detector. Although small in scale, our dataset provides a practical resource for developing and testing the real-world effectiveness of adversarial patch detectors.

\section{Conclusion}
In this work, we proposed a novel diffusion-based adversarial patch generation method \textbf{BadPatch} to generate naturalistic and customized patches based on a reference image. BadPatch utilizes image-to-latent mapping, incomplete diffusion optimization, and target mask control to create stylized adversarial patches. Extensive experiments across multiple object detection models validate the effectiveness of BadPatch. With BadPatch, we also created a physical-world adversarial dataset, \textbf{AdvT-shirt-1K}, comprising 1,131 images captured in diverse scenes and conditions, which validates the feasibility of BadPatch in physical-world environments.
{
    \small
    \bibliographystyle{ieeenat_fullname}
    \bibliography{main}

\begin{thebibliography}{64}
\providecommand{\natexlab}[1]{#1}
\providecommand{\url}[1]{\texttt{#1}}
\expandafter\ifx\csname urlstyle\endcsname\relax
  \providecommand{\doi}[1]{doi: #1}\else
  \providecommand{\doi}{doi: \begingroup \urlstyle{rm}\Url}\fi

\bibitem[Achiam et~al.(2023)Achiam, Adler, Agarwal, Ahmad, Akkaya, Aleman, Almeida, Altenschmidt, Altman, Anadkat, et~al.]{achiam2023gpt}
Josh Achiam, Steven Adler, Sandhini Agarwal, Lama Ahmad, Ilge Akkaya, Florencia~Leoni Aleman, Diogo Almeida, Janko Altenschmidt, Sam Altman, Shyamal Anadkat, et~al.
\newblock Gpt-4 technical report.
\newblock \emph{arXiv preprint arXiv:2303.08774}, 2023.

\bibitem[Andriluka et~al.(2014)Andriluka, Pishchulin, Gehler, and Schiele]{andriluka20142d}
Mykhaylo Andriluka, Leonid Pishchulin, Peter Gehler, and Bernt Schiele.
\newblock 2d human pose estimation: New benchmark and state of the art analysis.
\newblock In \emph{CVPR}, pages 3686--3693, 2014.

\bibitem[Athalye et~al.(2018)Athalye, Engstrom, Ilyas, and Kwok]{athalye2018synthesizing}
Anish Athalye, Logan Engstrom, Andrew Ilyas, and Kevin Kwok.
\newblock Synthesizing robust adversarial examples.
\newblock In \emph{ICML}, pages 284--293. PMLR, 2018.

\bibitem[Avrahami et~al.(2022)Avrahami, Lischinski, and Fried]{avrahami2022blended}
Omri Avrahami, Dani Lischinski, and Ohad Fried.
\newblock Blended diffusion for text-driven editing of natural images.
\newblock In \emph{CVPR}, pages 18208--18218, 2022.

\bibitem[Bochkovskiy et~al.(2020)Bochkovskiy, Wang, and Liao]{bochkovskiy2020yolov4}
Alexey Bochkovskiy, Chien-Yao Wang, and Hong-Yuan~Mark Liao.
\newblock Yolov4: Optimal speed and accuracy of object detection.
\newblock \emph{arXiv preprint arXiv:2004.10934}, 2020.

\bibitem[Brown et~al.(2017)Brown, Man{\'e}, Roy, Abadi, and Gilmer]{brown2017adversarial}
Tom~B Brown, Dandelion Man{\'e}, Aurko Roy, Mart{\'\i}n Abadi, and Justin Gilmer.
\newblock Adversarial patch.
\newblock \emph{arXiv preprint arXiv:1712.09665}, 2017.

\bibitem[Carion et~al.(2020)Carion, Massa, Synnaeve, Usunier, Kirillov, and Zagoruyko]{carion2020end}
Nicolas Carion, Francisco Massa, Gabriel Synnaeve, Nicolas Usunier, Alexander Kirillov, and Sergey Zagoruyko.
\newblock End-to-end object detection with transformers.
\newblock In \emph{ECCV}, pages 213--229. Springer, 2020.

\bibitem[Carlini and Wagner(2017)]{carlini2017towards}
Nicholas Carlini and David Wagner.
\newblock Towards evaluating the robustness of neural networks.
\newblock In \emph{2017 ieee symposium on security and privacy (sp)}, pages 39--57. Ieee, 2017.

\bibitem[Chen et~al.(2015)Chen, Seff, Kornhauser, and Xiao]{chen2015deepdriving}
Chenyi Chen, Ari Seff, Alain Kornhauser, and Jianxiong Xiao.
\newblock Deepdriving: Learning affordance for direct perception in autonomous driving.
\newblock In \emph{ICCV}, pages 2722--2730, 2015.

\bibitem[Chen et~al.(2023)Chen, Liu, Jiang, and Yan]{chen2023natural}
Xianyi Chen, Fazhan Liu, Dong Jiang, and Kai Yan.
\newblock Natural adversarial patch generation method based on latent diffusion model.
\newblock \emph{arXiv preprint arXiv:2312.16401}, 2023.

\bibitem[Chen et~al.(2024)Chen, Li, Wu, Jiang, Ding, and Zhang]{chen2024content}
Zhaoyu Chen, Bo Li, Shuang Wu, Kaixun Jiang, Shouhong Ding, and Wenqiang Zhang.
\newblock Content-based unrestricted adversarial attack.
\newblock \emph{NeurIPS}, 36, 2024.

\bibitem[Dalal and Triggs(2005)]{dalal2005histograms}
Navneet Dalal and Bill Triggs.
\newblock Histograms of oriented gradients for human detection.
\newblock In \emph{CVPR}, pages 886--893. Ieee, 2005.

\bibitem[Dhariwal and Nichol(2021)]{dhariwal2021diffusion}
Prafulla Dhariwal and Alexander Nichol.
\newblock Diffusion models beat gans on image synthesis.
\newblock \emph{NeurIPS}, 34:\penalty0 8780--8794, 2021.

\bibitem[Doan et~al.(2022)Doan, Xue, Ma, Abbasnejad, and Ranasinghe]{doan2022tnt}
Bao~Gia Doan, Minhui Xue, Shiqing Ma, Ehsan Abbasnejad, and Damith~C Ranasinghe.
\newblock Tnt attacks! universal naturalistic adversarial patches against deep neural network systems.
\newblock \emph{IEEE TIFS}, 17:\penalty0 3816--3830, 2022.

\bibitem[Evtimov et~al.(2017)Evtimov, Eykholt, Fernandes, Kohno, Li, Prakash, Rahmati, and Song]{evtimov2017robust}
Ivan Evtimov, Kevin Eykholt, Earlence Fernandes, Tadayoshi Kohno, Bo Li, Atul Prakash, Amir Rahmati, and Dawn Song.
\newblock Robust physical-world attacks on machine learning models.
\newblock \emph{arXiv preprint arXiv:1707.08945}, 2\penalty0 (3):\penalty0 4, 2017.

\bibitem[Eykholt et~al.(2018)Eykholt, Evtimov, Fernandes, Li, Rahmati, Xiao, Prakash, Kohno, and Song]{eykholt2018robust}
Kevin Eykholt, Ivan Evtimov, Earlence Fernandes, Bo Li, Amir Rahmati, Chaowei Xiao, Atul Prakash, Tadayoshi Kohno, and Dawn Song.
\newblock Robust physical-world attacks on deep learning visual classification.
\newblock In \emph{CVPR}, pages 1625--1634, 2018.

\bibitem[Goodfellow et~al.(2014{\natexlab{a}})Goodfellow, Pouget-Abadie, Mirza, Xu, Warde-Farley, Ozair, Courville, and Bengio]{goodfellow2014generative}
Ian Goodfellow, Jean Pouget-Abadie, Mehdi Mirza, Bing Xu, David Warde-Farley, Sherjil Ozair, Aaron Courville, and Yoshua Bengio.
\newblock Generative adversarial nets.
\newblock \emph{NeurIPS}, 27, 2014{\natexlab{a}}.

\bibitem[Goodfellow et~al.(2014{\natexlab{b}})Goodfellow, Shlens, and Szegedy]{goodfellow2014explaining}
Ian~J Goodfellow, Jonathon Shlens, and Christian Szegedy.
\newblock Explaining and harnessing adversarial examples.
\newblock \emph{arXiv preprint arXiv:1412.6572}, 2014{\natexlab{b}}.

\bibitem[Guesmi et~al.(2024{\natexlab{a}})Guesmi, Bilasco, Shafique, and Alouani]{guesmi2024advart}
Amira Guesmi, Ioan~Marius Bilasco, Muhammad Shafique, and Ihsen Alouani.
\newblock Advart: Adversarial art for camouflaged object detection attacks.
\newblock In \emph{ICIP}, pages 666--672. IEEE, 2024{\natexlab{a}}.

\bibitem[Guesmi et~al.(2024{\natexlab{b}})Guesmi, Ding, Hanif, Alouani, and Shafique]{guesmi2024dap}
Amira Guesmi, Ruitian Ding, Muhammad~Abdullah Hanif, Ihsen Alouani, and Muhammad Shafique.
\newblock Dap: A dynamic adversarial patch for evading person detectors.
\newblock In \emph{ICCV}, pages 24595--24604, 2024{\natexlab{b}}.

\bibitem[Hertz et~al.(2022)Hertz, Mokady, Tenenbaum, Aberman, Pritch, and Cohen-Or]{hertz2022prompt}
Amir Hertz, Ron Mokady, Jay Tenenbaum, Kfir Aberman, Yael Pritch, and Daniel Cohen-Or.
\newblock Prompt-to-prompt image editing with cross attention control.
\newblock \emph{arXiv preprint arXiv:2208.01626}, 2022.

\bibitem[Ho and Salimans(2022)]{ho2022classifier}
Jonathan Ho and Tim Salimans.
\newblock Classifier-free diffusion guidance.
\newblock \emph{arXiv preprint arXiv:2207.12598}, 2022.

\bibitem[Ho et~al.(2020)Ho, Jain, and Abbeel]{ho2020denoising}
Jonathan Ho, Ajay Jain, and Pieter Abbeel.
\newblock Denoising diffusion probabilistic models.
\newblock \emph{NeurIPS}, 33:\penalty0 6840--6851, 2020.

\bibitem[Hu et~al.(2021)Hu, Kung, Tan, Chen, Hua, and Cheng]{hu2021naturalistic}
Yu-Chih-Tuan Hu, Bo-Han Kung, Daniel~Stanley Tan, Jun-Cheng Chen, Kai-Lung Hua, and Wen-Huang Cheng.
\newblock Naturalistic physical adversarial patch for object detectors.
\newblock In \emph{CVPR}, pages 7848--7857, 2021.

\bibitem[Huang et~al.(2023)Huang, Chen, Chen, Wang, and Zhang]{huang2023t}
Hao Huang, Ziyan Chen, Huanran Chen, Yongtao Wang, and Kevin Zhang.
\newblock T-sea: Transfer-based self-ensemble attack on object detection.
\newblock In \emph{CVPR}, pages 20514--20523, 2023.

\bibitem[Huang et~al.(2020)Huang, Gao, Zhou, Xie, Yuille, Zou, and Liu]{huang2020universal}
Lifeng Huang, Chengying Gao, Yuyin Zhou, Cihang Xie, Alan~L Yuille, Changqing Zou, and Ning Liu.
\newblock Universal physical camouflage attacks on object detectors.
\newblock In \emph{CVPR}, pages 720--729, 2020.

\bibitem[Jan et~al.(2019)Jan, Messou, Lin, Huang, and Wang]{jan2019connecting}
Steve~TK Jan, Joseph Messou, Yen-Chen Lin, Jia-Bin Huang, and Gang Wang.
\newblock Connecting the digital and physical world: Improving the robustness of adversarial attacks.
\newblock In \emph{AAAI}, pages 962--969, 2019.

\bibitem[Jocher et~al.(2020)Jocher, Stoken, Borovec, NanoCode012, ChristopherSTAN, Changyu, Laughing, tkianai, Hogan, lorenzomammana, yxNONG, AlexWang1900, Diaconu, Marc, wanghaoyang0106, ml5ah, Doug, Ingham, Frederik, Guilhen, Hatovix, Poznanski, Fang, Yu, changyu98, Wang, Gupta, Akhtar, PetrDvoracek, and Rai]{glenn_jocher_2020_4154370}
Glenn Jocher, Alex Stoken, Jirka Borovec, NanoCode012, ChristopherSTAN, Liu Changyu, Laughing, tkianai, Adam Hogan, lorenzomammana, yxNONG, AlexWang1900, Laurentiu Diaconu, Marc, wanghaoyang0106, ml5ah, Doug, Francisco Ingham, Frederik, Guilhen, Hatovix, Jake Poznanski, Jiacong Fang, Lijun Yu, changyu98, Mingyu Wang, Naman Gupta, Osama Akhtar, PetrDvoracek, and Prashant Rai.
\newblock {ultralytics/yolov5}, 2020.

\bibitem[Kingma and Ba(2014)]{kingma2014adam}
Diederik~P Kingma and Jimmy Ba.
\newblock Adam: A method for stochastic optimization.
\newblock \emph{arXiv preprint arXiv:1412.6980}, 2014.

\bibitem[Kurakin et~al.(2018)Kurakin, Goodfellow, and Bengio]{kurakin2018adversarial}
Alexey Kurakin, Ian~J Goodfellow, and Samy Bengio.
\newblock Adversarial examples in the physical world.
\newblock In \emph{Artificial intelligence safety and security}, pages 99--112. Chapman and Hall/CRC, 2018.

\bibitem[Lapid and Sipper(2023)]{lapid2023patch}
Raz Lapid and Moshe Sipper.
\newblock Patch of invisibility: Naturalistic black-box adversarial attacks on object de-tectors.
\newblock \emph{arXiv preprint arXiv:2303.04238}, 2023.

\bibitem[Li et~al.(2024)Li, Liu, Yan, and Su]{li2024capgen}
Chaoqun Li, Zhuodong Liu, Huanqian Yan, and Hang Su.
\newblock Capgen: An environment-adaptive generator of adversarial patches.
\newblock \emph{arXiv preprint arXiv:2412.07253}, 2024.

\bibitem[Li et~al.(2022)Li, Li, Xiong, and Hoi]{li2022blip}
Junnan Li, Dongxu Li, Caiming Xiong, and Steven Hoi.
\newblock Blip: Bootstrapping language-image pre-training for unified vision-language understanding and generation.
\newblock In \emph{ICML}, pages 12888--12900. PMLR, 2022.

\bibitem[Lin et~al.(2023)Lin, Chu, Lin, Chen, and Wang]{lin2023diffusion}
Shuo-Yen Lin, Ernie Chu, Che-Hsien Lin, Jun-Cheng Chen, and Jia-Ching Wang.
\newblock Diffusion to confusion: Naturalistic adversarial patch generation based on diffusion model for object detector.
\newblock \emph{arXiv preprint arXiv:2307.08076}, 2023.

\bibitem[Lin et~al.(2014)Lin, Maire, Belongie, Hays, Perona, Ramanan, Doll{\'a}r, and Zitnick]{lin2014microsoft}
Tsung-Yi Lin, Michael Maire, Serge Belongie, James Hays, Pietro Perona, Deva Ramanan, Piotr Doll{\'a}r, and C~Lawrence Zitnick.
\newblock Microsoft coco: Common objects in context.
\newblock In \emph{ECCV}, pages 740--755. Springer, 2014.

\bibitem[Liu et~al.(2025)Liu, Wu, Wang, Han, Guo, Xiang, and Zhang]{liu2025beware}
Hangcheng Liu, Zhenhu Wu, Hao Wang, Xingshuo Han, Shangwei Guo, Tao Xiang, and Tianwei Zhang.
\newblock Beware of road markings: A new adversarial patch attack to monocular depth estimation.
\newblock \emph{NeurIPS}, 37:\penalty0 67689--67711, 2025.

\bibitem[Liu et~al.(2022)Liu, Levine, Lau, Chellappa, and Feizi]{liu2022segment}
Jiang Liu, Alexander Levine, Chun~Pong Lau, Rama Chellappa, and Soheil Feizi.
\newblock Segment and complete: Defending object detectors against adversarial patch attacks with robust patch detection.
\newblock In \emph{CVPR}, pages 14973--14982, 2022.

\bibitem[Ma et~al.(2025)Ma, Gao, Wang, Wang, Wang, Sun, Ding, Xu, Chen, Zhao, et~al.]{ma2025safety}
Xingjun Ma, Yifeng Gao, Yixu Wang, Ruofan Wang, Xin Wang, Ye Sun, Yifan Ding, Hengyuan Xu, Yunhao Chen, Yunhan Zhao, et~al.
\newblock Safety at scale: A comprehensive survey of large model safety.
\newblock \emph{arXiv preprint arXiv:2502.05206}, 2025.

\bibitem[Madry et~al.(2018)Madry, Makelov, Schmidt, Tsipras, and Vladu]{madry2018towards}
Aleksander Madry, Aleksandar Makelov, Ludwig Schmidt, Dimitris Tsipras, and Adrian Vladu.
\newblock Towards deep learning models resistant to adversarial attacks.
\newblock In \emph{ICLR}, 2018.

\bibitem[Miotto et~al.(2018)Miotto, Wang, Wang, Jiang, and Dudley]{miotto2018deep}
Riccardo Miotto, Fei Wang, Shuang Wang, Xiaoqian Jiang, and Joel~T Dudley.
\newblock Deep learning for healthcare: review, opportunities and challenges.
\newblock \emph{Briefings in bioinformatics}, 19\penalty0 (6):\penalty0 1236--1246, 2018.

\bibitem[Mokady et~al.(2023)Mokady, Hertz, Aberman, Pritch, and Cohen-Or]{mokady2023null}
Ron Mokady, Amir Hertz, Kfir Aberman, Yael Pritch, and Daniel Cohen-Or.
\newblock Null-text inversion for editing real images using guided diffusion models.
\newblock In \emph{CVPR}, pages 6038--6047, 2023.

\bibitem[Radford et~al.(2021)Radford, Kim, Hallacy, Ramesh, Goh, Agarwal, Sastry, Askell, Mishkin, Clark, et~al.]{radford2021learning}
Alec Radford, Jong~Wook Kim, Chris Hallacy, Aditya Ramesh, Gabriel Goh, Sandhini Agarwal, Girish Sastry, Amanda Askell, Pamela Mishkin, Jack Clark, et~al.
\newblock Learning transferable visual models from natural language supervision.
\newblock In \emph{ICML}, pages 8748--8763. PMLR, 2021.

\bibitem[Ramesh et~al.(2022)Ramesh, Dhariwal, Nichol, Chu, and Chen]{ramesh2022hierarchical}
Aditya Ramesh, Prafulla Dhariwal, Alex Nichol, Casey Chu, and Mark Chen.
\newblock Hierarchical text-conditional image generation with clip latents.
\newblock \emph{arXiv preprint arXiv:2204.06125}, 1\penalty0 (2):\penalty0 3, 2022.

\bibitem[Redmon and Farhadi(2018)]{redmon2018yolov3}
Joseph Redmon and Ali Farhadi.
\newblock Yolov3: An incremental improvement.
\newblock \emph{arXiv preprint arXiv:1804.02767}, 2018.

\bibitem[Ren et~al.(2015)Ren, He, Girshick, and Sun]{ren2015faster}
Shaoqing Ren, Kaiming He, Ross Girshick, and Jian Sun.
\newblock Faster r-cnn: Towards real-time object detection with region proposal networks.
\newblock \emph{NeurIPS}, 28, 2015.

\bibitem[Rombach et~al.(2022)Rombach, Blattmann, Lorenz, Esser, and Ommer]{rombach2022high}
Robin Rombach, Andreas Blattmann, Dominik Lorenz, Patrick Esser, and Bj{\"o}rn Ommer.
\newblock High-resolution image synthesis with latent diffusion models.
\newblock In \emph{CVPR}, pages 10684--10695, 2022.

\bibitem[Saharia et~al.(2022)Saharia, Chan, Saxena, Li, Whang, Denton, Ghasemipour, Gontijo~Lopes, Karagol~Ayan, Salimans, et~al.]{saharia2022photorealistic}
Chitwan Saharia, William Chan, Saurabh Saxena, Lala Li, Jay Whang, Emily~L Denton, Kamyar Ghasemipour, Raphael Gontijo~Lopes, Burcu Karagol~Ayan, Tim Salimans, et~al.
\newblock Photorealistic text-to-image diffusion models with deep language understanding.
\newblock \emph{NeurIPS}, 35:\penalty0 36479--36494, 2022.

\bibitem[Sharif et~al.(2016)Sharif, Bhagavatula, Bauer, and Reiter]{sharif2016accessorize}
Mahmood Sharif, Sruti Bhagavatula, Lujo Bauer, and Michael~K Reiter.
\newblock Accessorize to a crime: Real and stealthy attacks on state-of-the-art face recognition.
\newblock In \emph{ACM CCS}, pages 1528--1540, 2016.

\bibitem[Shuai et~al.(2024)Shuai, Ding, Ma, Tu, Jiang, and Tao]{shuai2024survey}
Xincheng Shuai, Henghui Ding, Xingjun Ma, Rongcheng Tu, Yu-Gang Jiang, and Dacheng Tao.
\newblock A survey of multimodal-guided image editing with text-to-image diffusion models.
\newblock \emph{arXiv preprint arXiv:2406.14555}, 2024.

\bibitem[Song et~al.(2020)Song, Meng, and Ermon]{song2020denoising}
Jiaming Song, Chenlin Meng, and Stefano Ermon.
\newblock Denoising diffusion implicit models.
\newblock \emph{arXiv preprint arXiv:2010.02502}, 2020.

\bibitem[Thys et~al.(2019)Thys, Van~Ranst, and Goedem{\'e}]{thys2019fooling}
Simen Thys, Wiebe Van~Ranst, and Toon Goedem{\'e}.
\newblock Fooling automated surveillance cameras: adversarial patches to attack person detection.
\newblock In \emph{CVPRW}, pages 0--0, 2019.

\bibitem[Tian et~al.(2025)Tian, Ye, and Doermann]{tian2025yolov12}
Yunjie Tian, Qixiang Ye, and David Doermann.
\newblock Yolov12: Attention-centric real-time object detectors.
\newblock \emph{arXiv preprint arXiv:2502.12524}, 2025.

\bibitem[Wang et~al.(2025{\natexlab{a}})Wang, Chen, Liu, Chen, Lin, Han, et~al.]{wang2025yolov10}
Ao Wang, Hui Chen, Lihao Liu, Kai Chen, Zijia Lin, Jungong Han, et~al.
\newblock Yolov10: Real-time end-to-end object detection.
\newblock \emph{NeurIPS}, 37:\penalty0 107984--108011, 2025{\natexlab{a}}.

\bibitem[Wang et~al.(2023)Wang, Bochkovskiy, and Liao]{wang2023yolov7}
Chien-Yao Wang, Alexey Bochkovskiy, and Hong-Yuan~Mark Liao.
\newblock Yolov7: Trainable bag-of-freebies sets new state-of-the-art for real-time object detectors.
\newblock In \emph{CVPR}, pages 7464--7475, 2023.

\bibitem[Wang et~al.(2025{\natexlab{b}})Wang, Zhang, Zhou, Wu, Yu, Zhao, Yin, Fu, Yan, Luo, et~al.]{wang2025comprehensive}
Kun Wang, Guibin Zhang, Zhenhong Zhou, Jiahao Wu, Miao Yu, Shiqian Zhao, Chenlong Yin, Jinhu Fu, Yibo Yan, Hanjun Luo, et~al.
\newblock A comprehensive survey in llm (-agent) full stack safety: Data, training and deployment.
\newblock \emph{arXiv preprint arXiv:2504.15585}, 2025{\natexlab{b}}.

\bibitem[Wei et~al.(2025)Wei, Wang, Zhang, Hou, Liu, Tang, and Wang]{wei2025revisiting}
Hui Wei, Zhixiang Wang, Kewei Zhang, Jiaqi Hou, Yuanwei Liu, Hao Tang, and Zheng Wang.
\newblock Revisiting adversarial patches for designing camera-agnostic attacks against person detection.
\newblock \emph{NeurIPS}, 37:\penalty0 8047--8064, 2025.

\bibitem[Wei et~al.(2018)Wei, Liang, Chen, and Cao]{wei2018transferable}
Xingxing Wei, Siyuan Liang, Ning Chen, and Xiaochun Cao.
\newblock Transferable adversarial attacks for image and video object detection.
\newblock \emph{arXiv preprint arXiv:1811.12641}, 2018.

\bibitem[Wu et~al.(2020)Wu, Lim, Davis, and Goldstein]{wu2020making}
Zuxuan Wu, Ser-Nam Lim, Larry~S Davis, and Tom Goldstein.
\newblock Making an invisibility cloak: Real world adversarial attacks on object detectors.
\newblock In \emph{ECCV}, pages 1--17. Springer, 2020.

\bibitem[Xie et~al.(2017)Xie, Wang, Zhang, Zhou, Xie, and Yuille]{xie2017adversarial}
Cihang Xie, Jianyu Wang, Zhishuai Zhang, Yuyin Zhou, Lingxi Xie, and Alan Yuille.
\newblock Adversarial examples for semantic segmentation and object detection.
\newblock In \emph{ICCV}, pages 1369--1378, 2017.

\bibitem[Xu et~al.(2020)Xu, Zhang, Liu, Fan, Sun, Chen, Chen, Wang, and Lin]{xu2020adversarial}
Kaidi Xu, Gaoyuan Zhang, Sijia Liu, Quanfu Fan, Mengshu Sun, Hongge Chen, Pin-Yu Chen, Yanzhi Wang, and Xue Lin.
\newblock Adversarial t-shirt! evading person detectors in a physical world.
\newblock In \emph{ECCV}, pages 665--681. Springer, 2020.

\bibitem[Xue et~al.(2024)Xue, Araujo, Hu, and Chen]{xue2024diffusion}
Haotian Xue, Alexandre Araujo, Bin Hu, and Yongxin Chen.
\newblock Diffusion-based adversarial sample generation for improved stealthiness and controllability.
\newblock \emph{NeurIPS}, 36, 2024.

\bibitem[Zhao et~al.(2024)Zhao, Lv, Xu, Wei, Wang, Dang, Liu, and Chen]{zhao2024detrs}
Yian Zhao, Wenyu Lv, Shangliang Xu, Jinman Wei, Guanzhong Wang, Qingqing Dang, Yi Liu, and Jie Chen.
\newblock Detrs beat yolos on real-time object detection.
\newblock In \emph{CVPR}, pages 16965--16974, 2024.

\bibitem[Zhong et~al.(2022)Zhong, Liu, Zhai, Jiang, and Ji]{zhong2022shadows}
Yiqi Zhong, Xianming Liu, Deming Zhai, Junjun Jiang, and Xiangyang Ji.
\newblock Shadows can be dangerous: Stealthy and effective physical-world adversarial attack by natural phenomenon.
\newblock In \emph{CVPR}, pages 15345--15354, 2022.

\bibitem[Zhu et~al.(2021)Zhu, Li, Li, Wang, and Hu]{zhu2021fooling}
Xiaopei Zhu, Xiao Li, Jianmin Li, Zheyao Wang, and Xiaolin Hu.
\newblock Fooling thermal infrared pedestrian detectors in real world using small bulbs.
\newblock In \emph{AAAI}, pages 3616--3624, 2021.

\end{thebibliography}
}

\end{document}